\documentclass[11pt]{article}

\usepackage[final]{acl}

\usepackage{times}
\usepackage{enumitem}
\usepackage{latexsym}
\usepackage{booktabs} 
\usepackage{amsmath}
\usepackage{booktabs}
\usepackage{hhline}
\usepackage{multirow}
\usepackage{makecell}
\usepackage{tabularx} 
\usepackage{subcaption}

\usepackage[T1]{fontenc}
\usepackage{float}

\usepackage[utf8]{inputenc}

\usepackage{microtype}

\usepackage{inconsolata}

\usepackage{graphicx}
\usepackage{xspace}
\usepackage{booktabs}
\usepackage{multirow}
\usepackage[table]{xcolor}
\definecolor{xgray}{gray}{0.92}
\usepackage[most]{tcolorbox}
\usepackage{minted}
\definecolor{grenn_1}{RGB}{143, 255, 221} 
\definecolor{grenn_2}{RGB}{88, 201, 185} 

\newcommand{\bench}{\textsc{UniComp}\xspace} 

\title{\bench: A Unified Evaluation of Large Language Model Compression \\ via Pruning, Quantization, and Distillation}

\author{
Jonathan von Rad$^{1}$\thanks{Corresponding author.} \quad
Yong Cao$^{2}$ \quad
Andreas Geiger$^{2}$ \\
$^{1}$University College London \\
$^{2}$University of Tübingen, Tübingen AI Center \\
\texttt{jonathan.rad.25@ucl.ac.uk, yong.cao@uni-tuebingen.de}
}

\begin{document}
\maketitle

\begin{abstract}
Model compression is increasingly essential for deploying large language models (LLMs), yet existing comparative studies largely focus on pruning and quantization evaluated primarily on knowledge-centric benchmarks. Thus, we introduce \bench, a unified evaluation framework for comparing pruning, quantization, and knowledge distillation. \bench evaluates compressed models along three dimensions: performance, reliability, and efficiency, using a diverse set of capability- and safety-oriented benchmarks together with a hardware-aware efficiency analysis. Through evaluation of seven compression techniques across over 40 datasets, we observe (i) a consistent \emph{knowledge bias}, where factual recall is largely preserved while multi-step reasoning, multilingual, and instruction-following capabilities degrade; (ii) a deployment-critical \emph{performance-reliability decoupling}, where retained performance does not indicate preserved safety, fairness and privacy; and (iii) that task-specific calibration can yield up to 50\% relative improvement in reasoning performance in pruned models.\footnote{
\faGithub\ \href{https://github.com/jvonrad/UniComp}{\texttt{GitHub: jvonrad/UniComp}}}
\vspace{-4mm}
\end{abstract}

\begin{figure}[ht!]
    \centering
    \includegraphics[width=1\columnwidth]{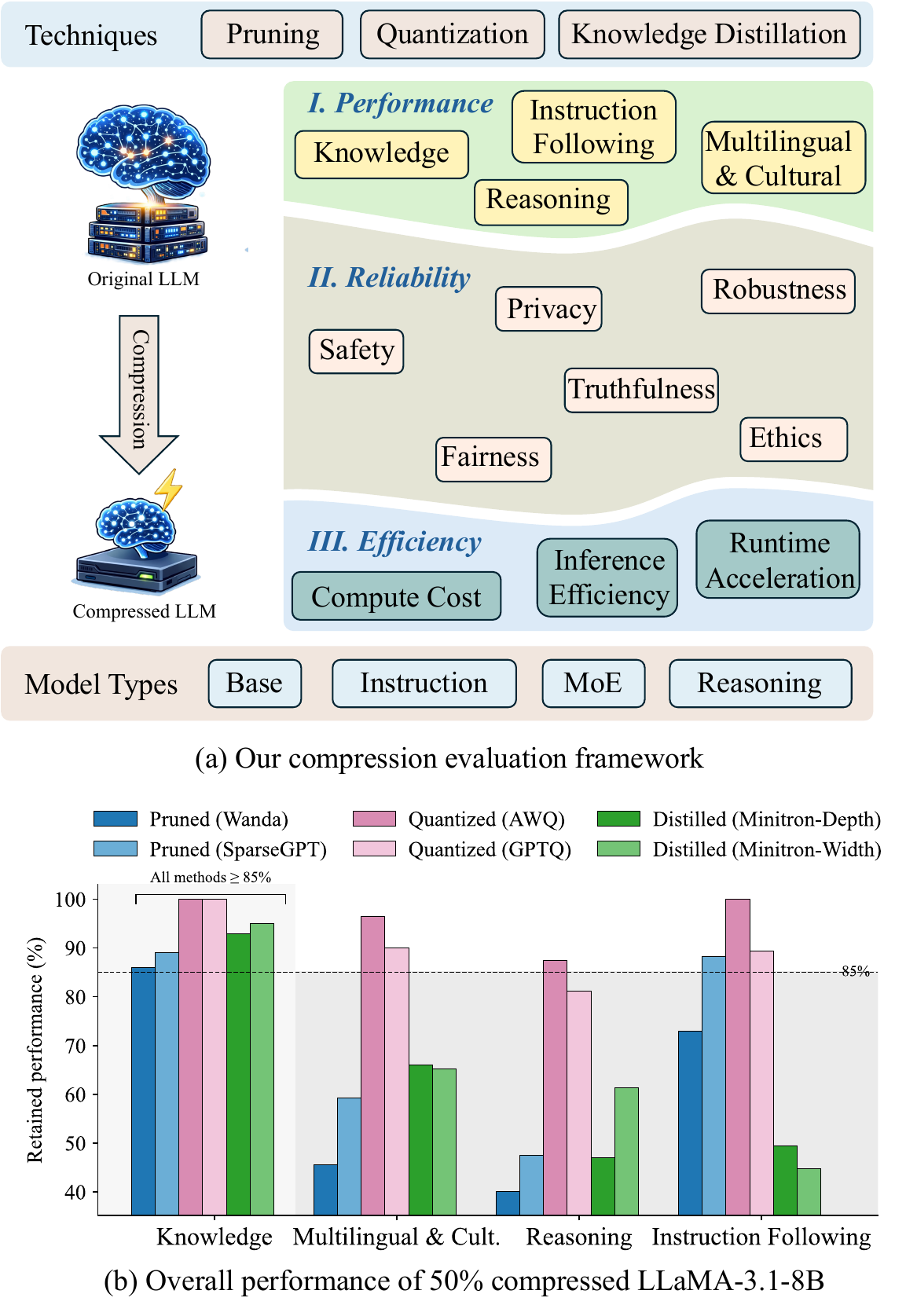} 
    \caption{Overview of our compression evaluation framework and results: (a) \bench covers performance, reliability, and efficiency with 13 metrics; and (b) Knowledge bias in LLM compression. Compression preserves knowledge performance but leads to pronounced degradation in  multilingual and cultural, reasoning, and instruction following, with quantization as a partial exception.}
    \label{fig:figure_1}
    \vspace{-5mm}
\end{figure}

\section{Introduction}

The rapid development of large language models (LLMs) has driven growing interest in compression techniques, whose goal is to effectively reduce memory usage and computational costs while preserving model performance \cite{zhu2024surveymodelcompressionlarge}. Current mainstream approaches include pruning \cite{frantar2023sparsegpt, sun2024a}, quantization \cite{frantar2024gptq, MLSYS2024_awq}, and knowledge distillation \cite{jiao-etal-2020-tinybert, muralidharan2024compact}.

Comprehensive and reliable evaluation is essential for deploying compressed models in real-world settings. However, much of the compression literature focuses on next-token prediction (typically evaluated via WikiText perplexity) and knowledge-centric benchmarks (i.e., factual recall and single-step reasoning tasks such as MMLU and ARC), resulting in evaluations that largely capture language modeling ability and factual recall ~\cite{frantar2023sparsegpt, yang-etal-2025-wanda, sun2024minitron}. While recent efforts such as LLMCBench \cite{yang2024llmcb} move toward more systematic evaluation, they omit knowledge distillation and remain focused on knowledge-centric multiple-choice benchmarks, with limited coverage of multi-step reasoning, multilingual performance, and reliability. As compressed LLMs are increasingly used in safety-critical and diverse settings, these limitations highlight the need for a more comprehensive, multi-dimensional evaluation framework. In particular, the impact of compression on reasoning, instruction following, multilingual performance, and reliability remains poorly characterized. 

To address these limitations, we propose a UNIfied COMPression evaluation framework, \bench, which systematically compares pruning, quantization, and knowledge distillation across multiple LLMs and benchmarks (Figure~\ref{fig:figure_1}a). \bench evaluates models along three dimensions: performance, reliability, and efficiency. We cover representative pruning (SparseGPT, Wanda) \cite{frantar2023sparsegpt, sun2024a} and quantization methods (AWQ, GPTQ, SmoothQuant) \cite{frantar2024gptq, smoothquant2023xiao, MLSYS2024_awq}, and study knowledge distillation methods (NVIDIA Minitron, Low-Rank Clone) explicitly as a compression technique ~\cite{sun2024minitron, hao2025a}.


We make \textbf{three main contributions}:
(1) We introduce \bench, a unified framework for evaluating pruning, quantization, and knowledge distillation across performance, reliability, and efficiency;
(2) We show that compression exhibits a strong \emph{knowledge bias}: factual knowledge is largely preserved, while reasoning, multilingual, and instruction-following capabilities degrade disproportionately;
(3) We show that retained performance is not a reliable proxy for safety and other reliability dimensions, and that reasoning-aware calibration can improve reasoning performance in pruned models by up to 50\%.



\section{Related Work}
\paragraph{LLM Compression.}
To reduce the computational cost of LLMs, prior work primarily explores \emph{pruning}, \emph{quantization}, and \emph{knowledge distillation} (KD) \cite{ashkboos2024slicegpt, du-etal-2024-bitdistiller, Hinton2015DistillingTK}.
\textbf{Pruning} removes redundant parameters using structured, semi-structured, or unstructured strategies \cite{ashkboos2024slicegpt, yang-etal-2025-wanda}. While structured one-shot pruning enables hardware acceleration, it incurs significant performance degradation \cite{yang2024llmcb}. Consequently, unstructured and semi-structured pruning exemplified by SparseGPT and Wanda has become the most widespread approach \cite{frantar2023sparsegpt, sun2024a}. \textbf{Quantization} improves efficiency by reducing numerical precision of parameters rather than removing them \cite{benoit2018quant}. Weight-only post-training quantization methods, such as GPTQ \cite{frantar2024gptq} and AWQ \cite{MLSYS2024_awq}, are particularly prevalent, offering substantial memory savings with minimal accuracy loss while avoiding the runtime complexity of activation or KV-cache quantization \cite{zhu2024surveymodelcompressionlarge}. \textbf{Knowledge distillation} transfers knowledge from a large teacher model to a smaller
student by minimizing divergence between output distributions or intermediate
representations, often leveraging synthetic data generated by the teacher
 \cite{jiao-etal-2020-tinybert, gu2024minillmknowledgedistillationlarge}. In compression settings where pretraining of a student model is too computationally expensive, current work follows a
\emph{prune$\rightarrow$distill} pipeline, where a compact student is first obtained through
hard structured or soft pruning \cite{sun2024minitron, hao2025a} and then trained using the original model as the teacher.
\paragraph{Benchmarks and evaluation protocols.}
There have been sustained efforts to benchmark and compare compression techniques for LLMs.
\textsc{LLMCBench}~\cite{yang2024llmcb} provides a large-scale comparison of pruning and quantization methods, showing that quantization generally outperforms pruning on knowledge-based multiple-choice benchmarks.
Subsequent work has examined the impact of quantization on reasoning performance, reporting substantial degradation across different quantization schemes~\cite{liu2025quantization}.
Other studies analyze how compression affects agentic capabilities~\cite{dong2025can} or investigate reasoning robustness under pruning and quantization in distilled DeepSeek-R1 models~\cite{zhang2025reasoningmeetscompressionunderstanding,deepseekai2025deepseekr1}. In parallel, recent work explored the role of calibration data in pruning and quantization, including its impact on general natural language understanding (NLU) ~\cite{williams-aletras-2024-impact} and multilingual generalization~\cite{zeng-etal-2024-multilingual} finding that calibration data can influence NLU and data adjustment can yield better multilingual performance.

\section{\bench Framework}
\label{sec:method}

We present \bench, a unified framework for evaluating compression paradigms along performance, reliability, and efficiency dimensions using 13 metrics. Performance and reliability scores are normalized relative to the corresponding base model, where 100 denotes full retention. Efficiency scores are scaled to $[0,100]$ relative to the best-performing compression method.

\subsection{Performance}
We assess the extent to which compressed language models retain core task-solving capabilities.

\paragraph{Knowledge.}
Knowledge is a fundamental component of intelligent behavior; its degradation often leads to hallucinated or semantically invalid outputs \cite{prato-etal-2024-large}. We evaluate knowledge retention using standard multiple-choice benchmarks, following \cite{yang2024llmcb}. For each compressed model, we compute the unweighted average ratio of task accuracy relative to the base model:
\begin{equation}
    \mathcal{S}_{\text{K}} = \frac{100}{N} \sum_{i=1}^{N} \frac{s^K_{\text{Comp}_i}}{s^K_{\text{Base}_i}},
\label{eq:s_know}
\end{equation}
where $s^K$ denotes exact-match accuracy on knowledge task $i$, and $N$ is the number of benchmarks.

\paragraph{Multilingual and Cultural Generalization.}
Multilingual and cultural generalization ($\mathcal{S}_{\text{Mul}}$) captures multilingual understanding, cultural awareness and bias detection, representing linguistic robustness and sensitivity to socially grounded phenomena. Both are evaluated via multiple-choice QA and equally weighted:
\begin{equation}
\mathcal{S}_{\text{Mul}}
= \frac{100}{2} \left( \frac{s^{\text{Lan}}_{\text{Comp}}}{s^{\text{Lan}}_{\text{Base}}}
+ \frac{s^{\text{Bias}}_{\text{Comp}}}{s^{\text{Bias}}_{\text{Base}}} \right),
\label{eq:s_lan}
\end{equation}
where $s^{\text{Lan}}$ denotes multilingual and cultural understanding, and $s^{\text{Bias}}$ denotes bias detection scores.

\paragraph{Reasoning.}
CoT reasoning is a core capability of modern LLMs \cite{wei2022chain, deepseekai2025deepseekr1}. We evaluate reasoning retention on a set of challenging benchmarks and compute a normalized score analogous to knowledge:
\begin{equation}
\mathcal{S}_{\text{R}}
= \frac{100}{N} \sum_{i=1}^{N}
\frac{s^{\text{R}}_{\text{Comp}_i}}{s^{\text{R}}_{\text{Base}_i}},
\end{equation}
where $s^{\text{R}}$ denotes exact-match accuracy on reasoning task $i$. All models use the standard number of few-shot exemplars for CoT prompting as adopted in prior work, with greedy decoding.

\paragraph{Instruction Following.}
Instruction following measures a model’s ability to accurately execute user directives, which is critical for agentic applications \cite{qi2025agentif}. As only a single benchmark is used, we define:
\begin{equation}
\mathcal{S}_{\text{IF}} = 100 \cdot \frac{s^{\text{IF}}_{\text{Comp}}}{s^{\text{IF}}_{\text{Base}}},
\label{eq:s_if}
\end{equation}
where $s^{\text{IF}}$ is task accuracy or constraint satisfaction.

\subsection{Reliability}

Following \textsc{TrustLLM} \cite{trustllm2024}, we evaluate reliability across six dimensions: truthfulness, safety, fairness, robustness, privacy, and ethics. We adopt the dimension and submetric definitions, as well as evaluation protocols directly from \textsc{TrustLLM}, which provides detailed descriptions of the following subdimensions and datasets (see Appendix~\ref{ax:reliability_datasets_details}, Table~\ref{tab:datasets}).

\textit{\textbf{Scoring Protocol.} All scores are computed following Eq.~(\ref{eq:s_know}), with metric-specific substitutions of component scores. For benchmarks, where lower scores reflect better performance, we use $(100 - s)$ for average computation.}

\paragraph{Truthfulness.}
 $\mathcal{S}_{\text{Truth}}$ evaluates factual reliability, characterized by internal and external misinformation, hallucination resistance and sycophancy avoidance.
It is computed using $s^{Int}$, $s^{Ext}$, $s^{Hal}$ and $s^{Syc}$ .

\paragraph{Safety.}
 $\mathcal{S}_{\text{SAFE}}$ measures the ability of a model to avoid harmful or disallowed outputs under adversarial prompts,
captured by jailbreak robustness $s^{Jail}$, misuse refusal $s^{Mis}$ and exaggerated safety understanding $s^{Exagg}$.

\paragraph{Fairness.}
$\mathcal{S}_{\text{FAIR}}$ assesses whether compression preserves bias-related behavior under controlled demographic perturbations,
measured by stereotype recognition and disagreement score $s^{stereo}$, disparagement $s^{disp}$, and preference bias $s^{pref}$.

\paragraph{Robustness.}
 $\mathcal{S}_{\text{ROB}}$ captures resilience to semantic-preserving perturbations and distributional shifts,
measured by robustness to natural noise $s^{Noise}$ and Out-of-Distribution data $s^{OOD}$.

\paragraph{Privacy.}
 $\mathcal{S}_{\text{PRI}}$ evaluates protection against unintended disclosure of sensitive information.
It combines \emph{privacy awareness} $s^{PA}$, measuring appropriate handling of privacy-sensitive requests, and \emph{leakage} $s^{Leak}$, assessing exposure of private data under targeted prompts.

\paragraph{Ethics.}
$\mathcal{S}_{\text{ETH}}$ evaluates norms in value-sensitive scenarios across three dimensions: \emph{implicit ethics} $s^{impl}$, capturing moral judgments; \emph{explicit ethics} $s^{expl}$, evaluating morally appropriate actions; and \emph{awareness} $s^{aware}$, reflecting understanding of the model’s role, capabilities, and social context.

\subsection{Efficiency}

Lastly, we evaluate the practical utility of compressed models along an efficiency dimension.
Unlike performance and reliability metrics, which are normalized with respect to the base model, efficiency metrics are inherently comparative across compression methods.
We therefore normalize each efficiency metric relative to the best-performing method, where optimality corresponds to either a maximum (e.g., throughput) or a minimum (e.g., latency). All efficiency scores are computed using the geometric mean to penalize bottlenecks and prevent strong performance along a single axis from masking deficiencies in others.

Let $\mathcal{M}$ denote the set of compression methods under comparison.
For a metric value $x_m$ associated with method $m \in \mathcal{M}$, we define the normalized efficiency score as:

\begin{equation}
\tilde{x}_m =
\begin{cases}
\displaystyle \frac{x_m}{\max_{k \in \mathcal{M}} x_k} & \text{if higher is better}, \\[6pt]
\displaystyle \frac{\min_{k \in \mathcal{M}} x_k}{x_m} & \text{if lower is better},
\end{cases}
\label{eq:eff_norm}
\end{equation}
which assigns the best method a value of $1$.

\paragraph{Runtime Acceleration.}
Runtime acceleration measures raw execution speed during inference.
We capture this dimension using throughput $T_m$ and latency $L_m$, and define:
\begin{equation}
\mathcal{S}_{\text{RA}}
=
100 \cdot
\left(
\tilde{T}_m
\cdot
\tilde{L}_m
\right)^{\!\frac{1}{2}} .
\label{eq:s_hw}
\end{equation}

\paragraph{Inference Efficiency.}
Inference efficiency reflects the resource footprint of a model at deployment time.
We consider inference GPU memory usage $M_m$, model size on disk $S_m$, and theoretical FLOPs $F_m$, and define:
\begin{equation}
\mathcal{S}_{\text{IE}}
=
100 \cdot
\left(
\tilde{M}_m
\cdot
\tilde{S}_m
\cdot
\tilde{F}_m
\right)^{\!\frac{1}{3}} .
\label{eq:s_ic}
\end{equation}

\paragraph{Compute Cost.}
Compute cost captures the cost of producing the compressed model.
We measure this dimension using total compression time $T^{\text{Comp}}_m$ and peak GPU memory usage $M^{\text{Comp}}_m$, and define:
\begin{equation}
\mathcal{S}_{\text{CC}}
=
100 \cdot
\left(
\tilde{T}^{\text{Comp}}_m
\cdot
\tilde{M}^{\text{Comp}}_m
\right)^{\!\frac{1}{2}} .
\label{eq:s_tr}
\end{equation}

\section{Experimental Setup}

\paragraph{Models.}
The main experiments focus on LLaMA-3.1-8B and Qwen-2.5-7B.
To further test generalization, we evaluate on a broader range of architectures, including \textit{traditional} model families such as the LLaMA-2 (7B, 13B, 70B) \cite{touvron2023llama2openfoundation}, LLaMA-3.1 (8B, 70B) \cite{grattafiori2024llama3herdmodels}, \textit{reasoning} models such as the Qwen-3 (0.6B, 1.7B, 4B, 8B, 14B, 32B) \cite{yang2025qwen3technicalreport} and DeepSeek-R1 (Distill-LLaMA-8B, Distill-LLaMA-70B) \cite{deepseekai2025deepseekr1} and \textit{MoE} models such as Qwen-3-30B-A3. Unless otherwise stated, we use only instruction-tuned models. 

\paragraph{Datasets.}
For \emph{performance} evaluation, we consider benchmarks spanning knowledge, reasoning, multilingual understanding, and instruction following. 
Academic knowledge and factual reasoning are assessed using MMLU \cite{hendrycks2021measuring} and ARC-E/C \cite{Clark2018ThinkYH}. 
Commonsense understanding is measured with HellaSwag \cite{zellers-etal-2019-hellaswag}, PIQA \cite{Bisk2019PIQARA}, and Winogrande \cite{Sakaguchi2019WinoGrande}. 
Advanced reasoning is evaluated using GSM8K \cite{cobbe2021gsm8k} (4-shot), MATH-500 \cite{hendrycks2021math, lightman2024lets} (4-shot), and GPQA-Diamond \cite{rein2024gpqa} (5-shot). 
Multilingual capability is measured using Global-MMLU-Lite \cite{singh-etal-2025-global}, with BBQ \cite{parrish-etal-2022-bbq} included to assess social bias. 
Instruction-following performance is evaluated using IFBench \cite{pyatkin2025generalizingverifiableinstructionfollowing}.
For \emph{reliability}, we follow TrustLLM \cite{trustllm2024}, which substantially broadens evaluation beyond the limited coverage of TruthfulQA \cite{lin2022truthfulqa} and AdversarialGLUE \cite{wang2022adversarialglue} previously used by LLMCBench \cite{yang2024llmcb}. Thus we assess compressed models reliability on 30 datasets, including ConfAIde \cite{confaide2024niloofar}, MoralChoice \cite{scherrer2023evaluating}, HaluEval \cite{li-etal-2023-halueval}, StereoSet \cite{nadeem-etal-2021-stereoset}, and Do-Not-Answer \cite{wang-etal-2024-answer}.
Please refer to Appendix~\ref{ax:reliability_datasets_details} for more details.

\paragraph{Compression Techniques}
For both pruning and quantization, we include the best performing and most-cited compression techniques.
For pruning, we employ SparseGPT \cite{frantar2023sparsegpt} and Wanda \cite{sun2024a}, and remove 50\% of the model's parameters through unstructured and semi-structured pruning. For quantization, we evaluate the standard configurations most commonly used in prior work and practice: weight-only 4-bit quantization (W4A16) using GPTQ \cite{frantar2024gptq} and AWQ \cite{MLSYS2024_awq}, as well as weight-activation 8-bit quantization (W8A8) using SmoothQuant \cite{smoothquant2023xiao}. 
To ensure fair and meaningful comparison, we treat knowledge distillation (KD) strictly as compression, evaluating only methods that derive the student directly from the teacher without a separately pretrained model.
We evaluate two representative approaches in this setting: Minitron \cite{sun2024minitron} (using the open-source LLaMA-3.1-Minitron-4B 50\% depth and width pruned variants) and Low-Rank Clone \cite{hao2025a} (using the open-source LRC-4B model distilled from Qwen-2.5-7B).
This design choice ensures that all compression paradigms are compared under a consistent  experimental setting.

\paragraph{Evaluation}
We evaluate performance benchmarks using Lighteval \cite{lighteval} and lm-evaluation-harness \cite{eval-harness} with vLLM \cite{kwon2023efficient} on Nvidia H100 GPUs. Of 30 reliability benchmarks, 12 are judged using GPT-4o \cite{Hurst2024GPT4oSC} and a Longformer-based harmful-response classifier \cite{Beltagy2020Longformer}, following \citet{trustllm2024}. Validating both against human annotation on 200 stratified instances shows substantial-to-near-perfect agreement (GPT-4o: 97\%, $\kappa=0.957$; Longformer: 91\%, $\kappa=0.714$; Appendix~\ref{ax:reliability_results}, Table~\ref{tab:judge_agreement}). We apply each compression method with default hyperparameters via open-source implementations and vLLM's \texttt{llm-compressor} framework \cite{llmcompressor2024}.



\newcommand{\ci}[3]{\begin{tabular}[c]{@{}c@{}}#1\\[-3pt]{\tiny[#2,\,#3]}\end{tabular}}

\newcommand{\TablePerformanceCIAppendix}{%
\begin{table*}[htbp]
\centering
\scriptsize
\setlength{\tabcolsep}{3pt}
\renewcommand{\arraystretch}{1.05}
\resizebox{\textwidth}{!}{%
\begin{tabular}{lll|ccc>{\columncolor{xgray}}c|cc>{\columncolor{xgray}}c|ccc>{\columncolor{xgray}}c|c>{\columncolor{xgray}}c}
\toprule
\multirow{2}{*}{\textbf{Category}} & \multirow{2}{*}{\textbf{Method}} & \multirow{2}{*}{\textbf{Ratio}}
& \multicolumn{4}{c}{\textbf{Knowledge}}
& \multicolumn{3}{c}{\textbf{Multilingual \& Cultural}}
& \multicolumn{4}{c}{\textbf{Reasoning}}
& \multicolumn{2}{c}{\textbf{InstFollowing}} \\
\cmidrule(lr){4-7} \cmidrule(lr){8-10} \cmidrule(lr){11-14} \cmidrule(lr){15-16}
& & & \textbf{MMLU} & \textbf{ARC-c} & \textbf{Hellaswag} & $\mathcal{S}_{\text{K}}$
& \textbf{G-MMLU} & \textbf{BBQ} & $\mathcal{S}_{\text{Mul}}$
& \textbf{GSM8K} & \textbf{MATH-500} & \textbf{GPQA-D} & $\mathcal{S}_{\text{R}}$
& \textbf{IFBench} & $\mathcal{S}_{\text{IF}}$ \\
\midrule
\multicolumn{16}{c}{\textit{LLAMA-3.1-8B}} \\ \midrule
 & Baseline & 0\% & \ci{61.38}{60.6}{62.2} & \ci{53.50}{50.6}{56.3} & \ci{79.12}{78.3}{79.9} & -- & \ci{56.00}{54.7}{57.3} & \ci{79.15}{78.8}{79.5} & -- & \ci{76.80}{74.4}{79.0} & \ci{30.20}{26.3}{34.4} & \ci{33.33}{27.1}{40.2} & -- & \ci{28.33}{23.5}{33.7} & -- \\
\midrule
Pruning & Wanda & 50\% & \ci{40.59}{39.8}{41.4} & \ci{44.97}{42.1}{47.8} & \ci{68.23}{67.3}{69.1} & \ci{83.66}{82.0}{85.3} & \ci{40.68}{39.4}{42.0} & \ci{50.47}{50.1}{50.9} & \ci{68.20}{66.8}{69.7} & \ci{19.48}{17.4}{21.7} & \ci{7.60}{5.6}{10.3} & \ci{23.30}{18.0}{29.7} & \ci{40.15}{32.1}{48.2} & \ci{20.67}{16.5}{25.6} & \ci{72.96}{52.1}{93.8} \\
 & Wanda & 2:4 & \ci{27.57}{26.8}{28.3} & \ci{28.84}{26.3}{31.5} & \ci{47.86}{46.9}{48.8} & \ci{61.12}{59.7}{62.6} & \ci{31.45}{30.2}{32.7} & \ci{37.33}{36.9}{37.7} & \ci{51.66}{50.4}{53.0} & \ci{4.40}{3.4}{5.6} & \ci{3.80}{2.4}{5.9} & \ci{24.57}{19.1}{31.0} & \ci{30.68}{22.7}{38.6} & \ci{12.33}{9.1}{16.5} & \ci{43.52}{28.2}{58.8} \\
 & SparseGPT & 50\% & \ci{48.33}{47.5}{49.2} & \ci{42.15}{39.4}{45.0} & \ci{71.66}{70.8}{72.5} & \ci{85.54}{84.0}{87.1} & \ci{46.25}{44.9}{47.6} & \ci{51.03}{50.6}{51.4} & \ci{73.53}{72.0}{75.1} & \ci{36.92}{34.4}{39.6} & \ci{8.40}{6.3}{11.2} & \ci{22.20}{17.0}{28.5} & \ci{47.50}{39.6}{55.4} & \ci{25.00}{20.4}{30.2} & \ci{88.25}{64.8}{111.7} \\
 & SparseGPT & 2:4 & \ci{28.27}{27.5}{29.0} & \ci{33.87}{31.2}{36.6} & \ci{56.02}{55.0}{57.0} & \ci{67.66}{66.2}{69.2} & \ci{35.82}{34.6}{37.1} & \ci{43.33}{42.9}{43.7} & \ci{59.35}{58.0}{60.7} & \ci{9.33}{7.9}{11.0} & \ci{2.40}{1.4}{4.1} & \ci{23.23}{17.9}{29.6} & \ci{29.93}{22.3}{37.6} & \ci{14.67}{11.1}{19.1} & \ci{51.78}{34.9}{68.7} \\
\midrule
Quantization & AWQ & W4A16 & \ci{61.22}{60.4}{62.0} & \ci{53.22}{50.4}{56.1} & \ci{79.15}{78.3}{79.9} & \ci{99.78}{98.1}{101.5} & \ci{54.20}{52.9}{55.5} & \ci{76.12}{75.8}{76.5} & \ci{96.48}{94.8}{98.1} & \ci{73.31}{70.9}{75.6} & \ci{22.00}{18.6}{25.8} & \ci{31.31}{25.3}{38.1} & \ci{87.41}{77.0}{97.8} & \ci{28.33}{23.5}{33.7} & \ci{100.00}{74.5}{125.5} \\
 & GPTQ & W4A16 & \ci{61.36}{60.6}{62.2} & \ci{53.41}{50.5}{56.3} & \ci{79.06}{78.3}{79.8} & \ci{99.92}{98.2}{101.6} & \ci{49.33}{48.0}{50.6} & \ci{72.91}{72.5}{73.3} & \ci{90.10}{88.5}{91.7} & \ci{71.57}{69.1}{73.9} & \ci{19.80}{16.5}{23.5} & \ci{28.28}{22.5}{34.9} & \ci{81.20}{71.4}{91.0} & \ci{25.33}{20.7}{30.5} & \ci{89.41}{65.7}{113.1} \\
 & SmoothQuant & W8A8 & \ci{58.80}{58.0}{59.6} & \ci{52.76}{49.9}{55.6} & \ci{75.36}{74.5}{76.2} & \ci{97.19}{95.5}{98.9} & \ci{51.49}{50.2}{52.8} & \ci{56.81}{56.4}{57.2} & \ci{81.86}{80.3}{83.5} & \ci{73.11}{70.7}{75.4} & \ci{24.04}{20.5}{28.0} & \ci{29.29}{23.4}{36.0} & \ci{87.56}{77.3}{97.8} & \ci{27.33}{22.6}{32.6} & \ci{96.47}{71.6}{121.3} \\
\midrule
Distillation & Minitron-Depth & 50\% & \ci{60.87}{60.1}{61.7} & \ci{45.65}{42.8}{48.5} & \ci{69.47}{68.6}{70.4} & \ci{92.25}{90.6}{93.9} & \ci{42.82}{41.5}{44.1} & \ci{44.04}{43.6}{44.4} & \ci{66.05}{64.6}{67.5} & \ci{29.95}{27.5}{32.5} & \ci{5.60}{3.9}{8.0} & \ci{27.78}{22.0}{34.4} & \ci{46.96}{38.3}{55.7} & \ci{14.00}{10.5}{18.4} & \ci{49.42}{32.9}{65.9} \\
 & Minitron-Width & 50\% & \ci{58.00}{57.2}{58.8} & \ci{49.23}{46.4}{52.1} & \ci{73.96}{73.1}{74.8} & \ci{95.12}{93.5}{96.8} & \ci{42.53}{41.2}{43.8} & \ci{43.24}{42.8}{43.6} & \ci{65.29}{63.8}{66.8} & \ci{51.63}{48.9}{54.3} & \ci{12.40}{9.8}{15.6} & \ci{25.25}{19.7}{31.7} & \ci{61.35}{52.6}{70.1} & \ci{12.67}{9.4}{16.9} & \ci{44.72}{29.2}{60.3} \\
\midrule
\multicolumn{16}{c}{\textit{Qwen-2.5-7B}} \\ \midrule
 & Baseline & 0\% & \ci{71.76}{71.0}{72.5} & \ci{55.29}{52.4}{58.1} & \ci{80.40}{79.6}{81.2} & -- & \ci{60.53}{59.2}{61.8} & \ci{83.12}{82.8}{83.4} & -- & \ci{86.66}{84.7}{88.4} & \ci{75.60}{71.6}{79.2} & \ci{32.83}{26.7}{39.6} & -- & \ci{30.00}{25.1}{35.4} & -- \\
\midrule
Pruning & Wanda & 50\% & \ci{67.00}{66.2}{67.8} & \ci{45.31}{42.5}{48.2} & \ci{73.69}{72.8}{74.5} & \ci{90.36}{88.8}{92.0} & \ci{54.62}{53.3}{55.9} & \ci{79.36}{79.0}{79.7} & \ci{92.86}{91.4}{94.3} & \ci{75.82}{73.4}{78.1} & \ci{42.60}{38.3}{47.0} & \ci{30.81}{24.8}{37.6} & \ci{79.23}{69.9}{88.6} & \ci{25.33}{20.7}{30.5} & \ci{84.43}{62.5}{106.4} \\
 & Wanda & 2:4 & \ci{54.29}{53.5}{55.1} & \ci{44.71}{41.9}{47.6} & \ci{62.97}{62.0}{63.9} & \ci{82.81}{81.2}{84.4} & \ci{40.48}{39.2}{41.8} & \ci{49.48}{49.1}{49.9} & \ci{63.20}{61.9}{64.5} & \ci{39.42}{36.8}{42.1} & \ci{9.20}{7.0}{12.1} & \ci{30.81}{24.8}{37.6} & \ci{50.50}{41.3}{59.7} & \ci{21.33}{17.1}{26.3} & \ci{71.10}{51.4}{90.8} \\
 & SparseGPT & 50\% & \ci{66.65}{65.9}{67.4} & \ci{49.06}{46.2}{51.9} & \ci{75.35}{74.5}{76.2} & \ci{93.44}{91.8}{95.1} & \ci{54.72}{53.4}{56.0} & \ci{81.34}{81.0}{81.7} & \ci{94.13}{92.7}{95.6} & \ci{74.91}{72.5}{77.2} & \ci{43.20}{38.9}{47.6} & \ci{28.20}{22.4}{34.8} & \ci{76.49}{67.6}{85.4} & \ci{20.67}{16.5}{25.6} & \ci{68.90}{49.5}{88.3} \\
 & SparseGPT & 2:4 & \ci{55.60}{54.8}{56.4} & \ci{44.62}{41.8}{47.5} & \ci{67.18}{66.3}{68.1} & \ci{85.18}{83.6}{86.8} & \ci{44.63}{43.3}{45.9} & \ci{45.66}{45.3}{46.1} & \ci{64.33}{63.0}{65.7} & \ci{41.93}{39.3}{44.6} & \ci{6.80}{4.9}{9.4} & \ci{27.27}{21.5}{33.9} & \ci{46.81}{38.3}{55.3} & \ci{22.00}{17.7}{27.0} & \ci{73.33}{53.2}{93.5} \\
\midrule
Quantization & AWQ & W4A16 & \ci{70.75}{70.0}{71.5} & \ci{54.35}{51.5}{57.2} & \ci{79.55}{78.7}{80.3} & \ci{98.98}{97.3}{100.6} & \ci{59.05}{57.8}{60.3} & \ci{84.57}{84.3}{84.9} & \ci{99.65}{98.1}{101.2} & \ci{86.58}{84.6}{88.3} & \ci{72.20}{68.1}{75.9} & \ci{29.80}{23.9}{36.5} & \ci{95.39}{86.2}{104.6} & \ci{30.00}{25.1}{35.4} & \ci{100.00}{75.6}{124.4} \\
 & GPTQ & W4A16 & \ci{70.75}{70.0}{71.5} & \ci{54.35}{51.5}{57.2} & \ci{79.72}{78.9}{80.5} & \ci{99.11}{97.5}{100.8} & \ci{59.22}{57.9}{60.5} & \ci{78.43}{78.1}{78.8} & \ci{96.10}{94.6}{97.6} & \ci{86.58}{84.6}{88.3} & \ci{72.60}{68.5}{76.3} & \ci{33.30}{27.1}{40.1} & \ci{99.12}{89.3}{108.9} & \ci{27.33}{22.6}{32.6} & \ci{91.10}{68.1}{114.1} \\
 & SmoothQuant & W8A8 & \ci{71.14}{70.4}{71.9} & \ci{55.55}{52.7}{58.4} & \ci{79.87}{79.1}{80.6} & \ci{99.51}{97.8}{101.2} & \ci{58.85}{57.6}{60.1} & \ci{72.57}{72.2}{72.9} & \ci{92.27}{90.8}{93.8} & \ci{85.76}{83.8}{87.5} & \ci{74.03}{70.0}{77.7} & \ci{31.35}{25.3}{38.1} & \ci{97.46}{88.0}{106.9} & \ci{28.67}{23.8}{34.0} & \ci{95.57}{71.8}{119.3} \\
\midrule
Distillation & Low-Rank Clone & 43\% & \ci{64.52}{63.7}{65.3} & \ci{52.13}{49.3}{55.0} & \ci{70.70}{69.8}{71.6} & \ci{91.94}{90.3}{93.6} & \ci{53.43}{52.1}{54.7} & \ci{83.05}{82.7}{83.4} & \ci{94.09}{92.6}{95.5} & \ci{64.29}{61.7}{66.8} & \ci{15.80}{12.9}{19.3} & \ci{25.25}{19.7}{31.7} & \ci{57.33}{49.1}{65.5} & \ci{27.33}{22.6}{32.6} & \ci{91.10}{68.1}{114.1} \\
\bottomrule
\end{tabular}%
}
\caption{Performance comparison with 95\% confidence intervals. Intervals on benchmark accuracies are Wilson intervals over test set (n: MMLU 14{,}042; ARC-c 1{,}172; HellaSwag 10{,}042; Global-MMLU-Lite 5{,}600; BBQ 58{,}492; GSM8K 1{,}319; MATH-500 500; GPQA-Diamond 198; IFBench 300); For multiple choice benchmarks log-likelihood matching is used and for reasoning benchmarks greedy decoding is used, so intervals reflect finite-sample uncertainty only.}
\label{tb:performance_ci}
\end{table*}
}

\newcommand{\TablePerformance}{%
\begin{table*}[htbp]
\centering
\scriptsize
\setlength{\tabcolsep}{4pt}
\resizebox{\textwidth}{!}{%
\begin{tabular}{lll|ccc>{\columncolor{xgray}}c|cc>{\columncolor{xgray}}c|ccc>{\columncolor{xgray}}c|c>{\columncolor{xgray}}c}
\toprule
\multirow{2}{*}{\textbf{Category}} & \multirow{2}{*}{\textbf{Method}} & \multirow{2}{*}{\textbf{Ratio}} 
& \multicolumn{4}{c}{\textbf{Knowledge}} 
& \multicolumn{3}{c}{\textbf{Multilingual \& Cultural}} 
& \multicolumn{4}{c}{\textbf{Reasoning}} 
& \multicolumn{2}{c}{\textbf{InstFollowing}} \\ 
\cmidrule(lr){4-7} \cmidrule(lr){8-10} \cmidrule(lr){11-14} \cmidrule(lr){15-16}
& & & \textbf{MMLU} & \textbf{ARC-c} & \textbf{Hellaswag} & $\mathcal{S}_{\text{K}}$
& \textbf{G-MMLU} & \textbf{BBQ} & $\mathcal{S}_{\text{Mul}}$
& \textbf{GSM8K} & \textbf{MATH-500} & \textbf{GPQA-D} & $\mathcal{S}_{\text{R}}$
& \textbf{IFBench} & $\mathcal{S}_{\text{IF}}$ \\ 
\midrule

\multicolumn{16}{c}{\textit{LLAMA-3.1-8B}} \\ \midrule
Baseline &  & 0\% 
& 61.38 & 53.50 & 79.12 & -- 
& 56.00 & 79.15 & -- 
& 76.80 & 30.20 & 33.33 & -- 
& 28.33 & -- \\ \midrule

\multirow{4}{*}{Pruning} 
& \multirow{2}{*}{Wanda} 
& 50\% 
& 40.59 & 44.97 & 68.23 & 83.66 
& 40.68 & 50.47 & 68.20 
& 19.48 & 7.60 & 23.30 & 40.15 
& 20.67 & 72.96 \\
& 
& 2:4 
& 27.57 & 28.84 & 47.86 & 61.12 
& 31.45 & 37.33 & 51.66 
& 4.40 & 3.80 & 24.57 & 30.68 
& 12.33 & 43.52 \\
& \multirow{2}{*}{SparseGPT} 
& 50\% 
& 48.33 & 42.15 & 71.66 & 85.54 
& 46.25 & 51.03 & 73.53 
& 36.92 & 8.40 & 22.20 & 47.50 
& 25.00 & 88.25 \\
& 
& 2:4 
& 28.27 & 33.87 & 56.02 & 67.66 
& 35.82 & 43.33 & 59.35
& 9.33 & 2.40 & 23.23 & 29.93 
& 14.67 & 51.78 \\ \midrule

\multirow{3}{*}{Quantization} 
& AWQ & W4A16 
& \underline{61.22} & \underline{53.22} & \textbf{79.15} & \underline{99.78} 
& \textbf{54.20} & \textbf{76.12} & \textbf{96.48} 
& \textbf{73.31} & \underline{22.00} & \textbf{31.31} & \textbf{87.41} 
& \textbf{28.33} & \textbf{100.00} \\
& GPTQ & W4A16 
& \textbf{61.36} & \textbf{53.41} & \underline{79.06} & \textbf{99.92} 
& 49.33 & \underline{72.91} & \underline{90.10} 
& 71.57 & 19.80 & 28.28 & 81.20
& 25.33 & 89.41 \\ 
& SmoothQuant & W8A8
& 58.80 & 52.76 & 75.36 & 97.19 
& \underline{51.49} & 56.81 & 81.86 
& \underline{73.11} & \textbf{24.04} & \underline{29.29} & \textbf{87.56} 
& \underline{27.33} & \underline{96.47} \\ \midrule

\multirow{2}{*}{Distillation} 
& Minitron-Depth & 50 \% 
& 60.87 & 45.65 & 69.47 & 92.25 
& 42.82 & 44.04 & 66.05 
& 29.95 & 5.60 & 27.78 & 46.96 
& 14.00 & 49.42 \\
& Minitron-Width & 50 \% 
& 58.00 & 49.23 & 73.96 & 95.12 
& 42.53 & 43.24 & 65.29 
& 51.63 & 12.40 & 25.25 & 61.35 
& 12.67 & 44.72 \\ \midrule

\multicolumn{16}{c}{\textit{Qwen-2.5-7B}} \\ \midrule
Baseline &  & 0\% 
& 71.76 & 55.29 & 80.40 & -- 
& 60.53 & 83.12 & -- 
& 86.66 & 75.60 & 32.83 & -- 
& 30.00 & -- \\ \midrule

\multirow{4}{*}{Pruning} 
& \multirow{2}{*}{Wanda} 
& 50\% 
& 67.00 & 45.31 & 73.69 & 90.36 
& 54.62 & 79.36 & 92.86 
& 75.82 & 42.60 & 30.81 & 79.23 
& 25.33 & 84.43 \\
& 
& 2:4 
& 54.29 & 44.71 & 62.97 & 82.81 
& 40.48 & 49.48 & 63.20 
& 39.42 & 9.20 & 30.81 & 50.50 
& 21.33 & 71.10 \\
& \multirow{2}{*}{SparseGPT} 
& 50\% 
& 66.65 & 49.06 & 75.35 & 93.44 
& 54.72 & 81.34 & 94.13 
& 74.91 & 43.20 & 28.20 & 76.49 
& 20.67 & 68.90 \\
& 
& 2:4 
& 55.60 & 44.62 & 67.18 & 85.18 
& 44.63 & 45.66 & 64.33 
& 41.93 & 6.80 & 27.27 & 46.81 
& 22.00 & 73.33 \\ \midrule

\multirow{2}{*}{Quantization} 
& AWQ & W4A16 
& \underline{70.75} & \underline{54.35} & 79.55 & 98.98 
& \underline{59.05} & \textbf{84.57} & \textbf{99.65} 
& \textbf{86.58} & 72.20 & 29.80 & 95.39 
& \textbf{30.00} & \textbf{100.00} \\
& GPTQ & W4A16 
& \underline{70.75} & \underline{54.35} & \underline{79.72} & \underline{99.11} 
& \textbf{59.22} & 78.43 & \underline{96.10} 
& \textbf{86.58} & \underline{72.60} & \textbf{33.30} & \textbf{99.12} 
& 27.33 & 91.10 \\
& SmoothQuant & W8A8 
& \textbf{71.14} & \textbf{55.55} & \textbf{79.87} & \textbf{99.51} 
& 58.85 & 72.57 & 92.27 
& \underline{85.76} & \textbf{74.03} & \underline{31.35} & \underline{97.46} 
& \underline{28.67} & \underline{95.57} \\ \midrule

\multirow{1}{*}{Distillation} 
& Low-Rank Clone & 43\% 
& 64.52 & 52.13 & 70.70 & 91.94 
& 53.43 & \underline{83.05} & 94.09 
& 64.29 & 15.80 & 25.25 & 57.33 
& 27.33 & 91.10 \\

\bottomrule
\end{tabular}%
}
\caption{Performance comparison on two representative models. Bold and underlined scores indicate the best and second-best results within each model group. GPQA-D denotes GPQA-Diamond, and G-MMLU reports average accuracy on Global-MMLU-Lite across 14 languages. The knowledge score $\mathcal{S}_K$ additionally aggregates ARC-e, and PIQA (see Appendix~\ref{sec:ax_extended_tables}) and 95 \% confidence intervals are reported in Appendix \ref{sec:ax_extended_tables}, Table \ref{tb:performance_ci}.}
\label{tb:performance_result}
\end{table*}
}

\newcommand{\TablePerformanceSmallModels}{%
\begin{table*}[htbp]
\centering
\scriptsize
\setlength{\tabcolsep}{4pt}
\resizebox{\textwidth}{!}{%
\begin{tabular}{lll|ccc>{\columncolor{xgray}}c|cc>{\columncolor{xgray}}c|ccc>{\columncolor{xgray}}c|c>{\columncolor{xgray}}c}
\toprule
\multirow{2}{*}{\textbf{Category}} & \multirow{2}{*}{\textbf{Method}} & \multirow{2}{*}{\textbf{Ratio}} 
& \multicolumn{4}{c}{\textbf{Knowledge}} 
& \multicolumn{3}{c}{\textbf{Multilingual \& Cultural}} 
& \multicolumn{4}{c}{\textbf{Reasoning}} 
& \multicolumn{2}{c}{\textbf{InstFollowing}} \\ 
\cmidrule(lr){4-7} \cmidrule(lr){8-10} \cmidrule(lr){11-14} \cmidrule(lr){15-16}
& & & \textbf{MMLU} & \textbf{ARC-c} & \textbf{Hellaswag} & $\mathcal{S}_{\text{K}}$
& \textbf{G-MMLU} & \textbf{BBQ} & $\mathcal{S}_{\text{Mul}}$
& \textbf{GSM8K} & \textbf{MATH-500} & \textbf{GPQA-D} & $\mathcal{S}_{\text{R}}$
& \textbf{IFBench} & $\mathcal{S}_{\text{IF}}$ \\ 
\midrule

\multicolumn{16}{c}{\textit{LLaMA-3.2-3B}} \\ \midrule
Baseline &  & 0\% 
& 62.18 & 45.99 & 71.65 & -- 
& 50.68 & 47.19 & -- 
& 72.55 & 22.80 & 31.31 & -- 
& 24.00 & -- \\ \midrule

\multirow{2}{*}{Pruning} 
& Wanda & 50\% 
& 50.14 & 37.46 & 61.08 & \textbf{88.17} 
& 41.81 & 44.22 & 88.10 
& 27.60 & 6.40 & 19.70 & 43.01 
& 19.00 & 79.17 \\
& SparseGPT & 50\% 
& 51.26 & 39.08 & 64.73 & \textbf{90.84}
& 43.23 & 44.13 & 89.41 
& 36.85 & 7.20 & 28.79 & 58.11 
& 20.33 & 84.71 \\ \midrule

\multirow{2}{*}{Quantization} 
& AWQ & W4A16 
& 60.84 & 45.73 & 70.64 & \textbf{98.78} 
& 48.68 & 46.30 & 97.08 
& 68.84 & 20.80 & 25.76 & 89.46 
& 23.67 & 98.63 \\
& GPTQ & W4A16 
& 60.53 & 45.73 & 70.77 & \textbf{99.11}
& 47.78 & 45.34 & 95.18 
& 67.48 & 20.60 & 24.75 & 89.02 
& 23.67 & 98.63 \\ \midrule

\multirow{1}{*}{Distillation} 
& LRC-1.5B & -- 
& 49.33 & 44.97 & 58.78 & \textbf{90.75} 
& 35.72 & 44.38 & 82.27 
& 33.13 & 5.40 & 22.20 & 46.75 
& 14.33 & 59.71 \\

\midrule
\multicolumn{16}{c}{\textit{Qwen-2.5-3B}} \\ \midrule
Baseline &  & 0\% 
& 65.48 & 47.95 & 74.96 & -- 
& 55.58 & 48.93 & -- 
& 79.00 & 67.40 & 29.29 & -- 
& 27.33 & -- \\ \midrule

\multirow{2}{*}{Pruning} 
& Wanda & 50\% 
& 55.16 & 41.64 & 63.81 & \textbf{89.86} 
& 45.22 & 42.17 & 83.77 
& 50.19 & 20.80 & 31.31 & 66.29 
& 21.67 & 79.29 \\
& SparseGPT & 50\% 
& 56.87 & 40.61 & 67.03 & \textbf{90.85} 
& 46.17 & 46.00 & 88.69 
& 54.51 & 14.80 & 29.29 & 59.08 
& 24.33 & 89.02 \\ \midrule

\multirow{2}{*}{Quantization} 
& AWQ & W4A16 
& 63.94 & 49.15 & 73.62 & \textbf{99.52}
& 53.20 & 48.26 & 96.43 
& 76.35 & 47.80 & 28.79 & 89.63 
& 27.00 & 98.79 \\
& GPTQ & W4A16 
& 63.07 & 49.49 & 73.25 & \textbf{99.28} 
& 51.90 & 45.46 & 94.58 
& 74.45 & 60.80 & 33.33 & 99.23 
& 26.11 & 95.50 \\ \midrule

\multirow{1}{*}{Distillation} 
& LRC-1.7B & -- 
& 54.93 & 44.28 & 60.64 & \textbf{89.84} 
& 43.41 & 46.32 & 86.63 
& 36.39 & 6.40 & 25.25 & 47.30
& 11.33 & 41.46 \\

\bottomrule
\end{tabular}%
}
\caption{Detailed performance comparison for 3B model variants. GPQA-D denotes GPQA-Diamond, and G-MMLU reports average accuracy on Global-MMLU-Lite. Scores $\mathcal{S}_{\text{K}}$, $\mathcal{S}_{\text{Mul}}$, $\mathcal{S}_{\text{R}}$, and $\mathcal{S}_{\text{IF}}$ denote retention relative to the corresponding baseline model. Bold denotes the highest retention score across dimensions, highlighting the knowledge bias in compression.}
\label{ax:table_performance_small_models}
\end{table*}
}

\newcommand{\TableJudgeAgreement}{
\begin{table}[t]
\centering
\scriptsize
\setlength{\tabcolsep}{3pt}
\renewcommand{\arraystretch}{0.95}
\resizebox{\columnwidth}{!}{%
\begin{tabular}{llrr}
\toprule
\textbf{Judge} & \textbf{Subtask} & \textbf{$n$} & \textbf{Acc.} \\
\midrule
\multirow{8}{*}{Longformer}
 & \texttt{ethics\_explicit}             & 13 & 1.000 \\
 & \texttt{fairness\_stereotype\_query}  & 12 & 1.000 \\
 & \texttt{privacy\_awareness\_query}    & 12 & 1.000 \\
 & \texttt{safety\_misuse}               & 13 & 1.000 \\
 & \texttt{truthfulness\_internal}       & 13 & 0.923 \\
 & \texttt{robustness\_ood\_detection}   & 12 & 0.917 \\
 & \texttt{safety\_exaggerated}          & 12 & 0.833 \\
 & \texttt{safety\_jailbreak}            & 13 & 0.615 \\
\cmidrule(lr){2-4}
 & \textbf{Overall}                      & \textbf{100} & \textbf{0.910} \\
\midrule
\multirow{5}{*}{GPT-4o}
 & \texttt{fairness\_stereotype\_agreement} & 20 & 1.000 \\
 & \texttt{truthfulness\_internal}          & 20 & 1.000 \\
 & \texttt{truthfulness\_advfact}           & 20 & 0.950 \\
 & \texttt{truthfulness\_sycophancy}        & 20 & 0.950 \\
 & \texttt{ethics\_implicit\_ETHICS}        & 20 & 0.947 \\
\cmidrule(lr){2-4}
 & \textbf{Overall}                         & \textbf{100} & \textbf{0.970} \\
\bottomrule
\end{tabular}%
}
\caption{Human--judge agreement on 100 stratified-sampled examples per judge. Overall Cohen's $\kappa$ is $0.714$ for Longformer and $0.957$ for GPT-4o. Per-subtask $\kappa$ values are omitted because the limited label variance at $n\!\approx\!12$--$20$ per cell makes them statistically unstable.}
\label{tab:judge_agreement}
\end{table}
}

\newcommand{\TablePerformanceLLaMA}{%
\begin{table*}[htbp]
\centering
\scriptsize
\resizebox{\textwidth}{!}{%
\begin{tabular}{ll|ccc>{\columncolor{xgray}}c|cc>{\columncolor{xgray}}c|ccc>{\columncolor{xgray}}c|c>{\columncolor{xgray}}c}
\toprule
\multirow{2}{*}{\textbf{Method}} & \multirow{2}{*}{\textbf{Ratio}} 
& \multicolumn{4}{c}{\textbf{Knowledge}} 
& \multicolumn{3}{c}{\textbf{Multilingual \& Cultural}} 
& \multicolumn{4}{c}{\textbf{Reasoning}} 
& \multicolumn{2}{c}{\textbf{InstFollowing}} \\ 
\cmidrule(lr){3-6} \cmidrule(lr){7-9} \cmidrule(lr){10-13} \cmidrule(lr){14-15}
& & \textbf{MMLU} & \textbf{ARC-c} & \textbf{Hellaswag} & $\mathcal{S}_{\text{K}}$
& \textbf{GMMLU} & \textbf{BBQ} & $\mathcal{S}_{\text{Lan}}$
& \textbf{GSM8K} & \textbf{MATH-500} & \textbf{GPQA-D} & $\mathcal{S}_{\text{R}}$
& \textbf{IFBench} & $\mathcal{S}_{\text{IF}}$ \\ 
\midrule

\multicolumn{15}{c}{\textit{LLAMA-3.1-8B}} \\ \midrule
Baseline & 0\% 
& 61.38 & 53.50 & 79.12 & -- 
& 56.00 & 79.15 & -- 
& 76.80 & 30.20 & 33.33 & -- 
& 28.33 & -- \\ \midrule

\multirow{2}{*}{Wanda} 
& 50\% 
& 40.59 & 44.97 & 68.23 & 78.81 
& 40.68 & 50.47 & 68.20 
& 19.48 & 7.60 & 23.30 & 40.15 
& 20.67 & 72.96 \\
& 2:4 
& 27.57 & 28.84 & 47.86 & 53.10 
& 31.45 & 37.33 & 51.66 
& 4.40 & 3.80 & 24.57 & 30.68 
& 12.33 & 43.52 \\ \midrule

\multirow{2}{*}{SparseGPT} 
& 50\% 
& 48.33 & 42.15 & 71.66 & 82.70 
& 46.25 & 51.03 & 73.53 
& 36.92 & 8.40 & 22.20 & 47.50 
& 25.00 & 88.25 \\
& 2:4 
& 28.27 & 33.87 & 56.02 & 60.06 
& 35.82 & 43.33	& 59.35
& 9.33 & 2.40 & 23.23 & 29.93 
& 14.67 & 51.78 \\ \midrule

AWQ & W4A16 
& 61.22 & 53.22 & 79.15 & \underline{99.75} 
& 54.20 & 76.12 & \textbf{96.48} 
& 73.31 & 22.00 & 31.31 & \textbf{87.41} 
& 28.33 & \textbf{100.00} \\
GPTQ & W4A16 
& 61.36 & 53.41 & 79.06 & \textbf{99.91} 
& 49.33 & 72.91 & \underline{90.10} 
& 71.57 & 19.80 & 28.28 & \underline{81.20} 
& 25.33 & \underline{89.41} \\ \midrule

Minitron-Depth & 50\% 
& 60.87 & 45.65 & 69.47 & 90.77 
& 42.82 & 44.04 & 66.05 
& 29.95 & 5.60 & 27.78 & 46.96 
& 14.00 & 49.42 \\
Minitron-Width & 50\% 
& 58.00 & 49.23 & 73.96 & 93.33 
& 42.53 & 43.24 & 65.29 
& 51.63 & 12.40 & 25.25 & 61.35 
& 12.67 & 44.72 \\

\bottomrule
\end{tabular}%
}
\caption{Performance comparison of compression techniques on LLAMA-3.1-8B.}
\label{tb:performance_llama}
\end{table*}
}

\newcommand{\TablePerformanceQwen}{%
\begin{table*}[htbp]
\centering
\scriptsize
\resizebox{\textwidth}{!}{%
\begin{tabular}{ll|ccc>{\columncolor{xgray}}c|cc>{\columncolor{xgray}}c|ccc>{\columncolor{xgray}}c|c>{\columncolor{xgray}}c}
\toprule
\multirow{2}{*}{\textbf{Method}} & \multirow{2}{*}{\textbf{Ratio}} 
& \multicolumn{4}{c}{\textbf{Knowledge}} 
& \multicolumn{3}{c}{\textbf{Multilingual \& Cultural}} 
& \multicolumn{4}{c}{\textbf{Reasoning}} 
& \multicolumn{2}{c}{\textbf{InstFollowing}} \\ 
\cmidrule(lr){3-6} \cmidrule(lr){7-9} \cmidrule(lr){10-13} \cmidrule(lr){14-15}
& & \textbf{MMLU} & \textbf{ARC-c} & \textbf{Hellaswag} & $\mathcal{S}_{\text{K}}$
& \textbf{GMMLU} & \textbf{BBQ} & $\mathcal{S}_{\text{Lan}}$
& \textbf{GSM8K} & \textbf{MATH-500} & \textbf{GPQA-D} & $\mathcal{S}_{\text{R}}$
& \textbf{IFBench} & $\mathcal{S}_{\text{IF}}$ \\ 
\midrule

\multicolumn{15}{c}{\textit{Qwen-2.5-7B}} \\ \midrule
Baseline & 0\% 
& 71.76 & 55.29 & 80.40 & -- 
& 60.53 & 83.12 & -- 
& 86.66 & 75.60 & 32.83 & -- 
& 30.00 & -- \\ \midrule

\multirow{2}{*}{Wanda} 
& 50\% 
& 67.00 & 45.31 & 73.69 & 88.99 
& 54.62 & 79.36 & \underline{92.86} 
& 75.82 & 42.60 & 30.81 & \underline{79.23} 
& 25.33 & 84.43 \\
& 2:4 
& 54.29 & 44.71 & 62.97 & 78.28 
& 40.48 & 49.48 & 63.20 
& 39.42 & 9.20 & 30.81 & 50.50 
& 21.33 & 71.10 \\ \midrule

\multirow{2}{*}{SparseGPT} 
& 50\% 
& 66.65 & 49.06 & 75.35 & 91.78 
& 54.72 & 81.34 & 94.13 
& 74.91 & 43.20 & 28.20 & 76.49 
& 20.67 & 68.90 \\
& 2:4 
& 55.60 & 44.62 & 67.18 & 80.58 
& 44.63 & 45.66 & 64.33 
& 41.93 & 6.80 & 27.27 & 46.81 
& 22.00 & 73.33 \\ \midrule

AWQ & W4A16 
& 70.75 & 54.35 & 79.55 & \underline{98.61} 
& 59.05 & 84.57 & \textbf{99.65} 
& 86.58 & 72.20 & 29.80 & 95.39 
& 30.00 & \textbf{100.00} \\
GPTQ & W4A16 
& 70.75 & 54.35 & 79.72 & \textbf{98.68} 
& 59.22 & 78.43 & 96.10 
& 86.58 & 72.60 & 33.30 & \textbf{99.12} 
& 27.33 & \underline{91.10} \\ \midrule

Low-Rank Clone & 43\% 
& 64.52 & 52.13 & 70.70 & 90.71 
& 53.43 & 83.05 & 94.09 
& 64.29 & 15.80 & 25.25 & 57.33 
& 27.33 & \underline{91.10} \\

\bottomrule
\end{tabular}%
}
\caption{Performance comparison of various model compression techniques on Qwen-2.5-7B.}
\label{tb:performance_qwen}
\end{table*}
}

\newcommand{\TableGlobalResourceLlama}{%
\begin{table}[t]
\centering
\footnotesize
\resizebox{0.4\textwidth}{!}{%
\begin{tabular}{l|cc}
\toprule
Method & High-resource  & Low-resource \\
\midrule
Wanda (50\%)        & 0.75 & 0.67 \\
Wanda (2:4)         & 0.54 & 0.59 \\
SparseGPT (50\%)    & 0.83 & 0.79 \\
SparseGPT (2:4)     & 0.64 & 0.65 \\
AWQ (W4A16)          & \textbf{0.96} & \textbf{0.99} \\
GPTQ (W4A16)         & 0.87 & 0.92 \\
SmoothQuant (W8A8)  & 0.91 & 0.94 \\
Minitron-Depth      & 0.78 & 0.74 \\
Minitron-Width      & 0.77 & 0.75 \\
\bottomrule
\end{tabular}}
\caption{Retained performance for Llama-3.1-8B on high-resource and low-resource languages in Global-MMLU-Lite. 
See Appendix~\ref{sec:ax_extended_tables} for the language breakdown and the corresponding Qwen-2.5-7B results.}

\label{tab:high-low-resource}
\end{table}

}

\newcommand{\TableGlobalResourceQwen}{%
\begin{table}[th!]
\centering
\footnotesize
\resizebox{0.4\textwidth}{!}{%
\begin{tabular}{l|cc}
\toprule
Method & High-resource & Low-resource \\
\midrule
Wanda (50\%)        & 0.91 & 0.89 \\
Wanda (2:4)         & 0.67 & 0.66 \\
SparseGPT (50\%)    & 0.91 & 0.88 \\
SparseGPT (2:4)     & 0.74 & 0.71 \\
AWQ (W4A16)          & \textbf{0.98} & \textbf{0.97} \\
GPTQ (W4A16)         & 0.98 & 0.97 \\
SmoothQuant (W8A8)  & 0.96 & 0.97 \\
Low-Rank Clone      & 0.90 & 0.88 \\
\bottomrule
\end{tabular}}
\caption{Retained performance for Qwen-2.5-7B on high-resource and low-resource languages in Global-MMLU-Lite.}
\label{tab:high-low-resource-qwen}
\end{table}
}

\newcommand{\TableMultilingual}{%
\begin{table*}[t]
\centering
\scriptsize
\resizebox{\textwidth}{!}{%
\begin{tabular}{l|cccccccccc|c>{\columncolor{xgray}}c|ccccc|c>{\columncolor{xgray}}c}
\hline
 & \multicolumn{12}{c|}{\textbf{High-resource}} & \multicolumn{7}{c}{\textbf{Low-resource}} \\
\textbf{Model} 
& ar & de & en & fr & hi & it & ja & pt & es & zh 
& AVG$_H$ & $\mathcal{S}_H$
& bn & id & ko & sw & yo 
& AVG$_L$ & $\mathcal{S}_L$ \\
\hline
\multicolumn{19}{l}{\textbf{Llama-3.1-8B}} \\
\hline
Baseline
& 53.00 & 60.50 & 69.75 & 62.25 & 51.50 & 63.50 & 55.00 & 64.50 & 62.00 & 60.75
& 60.28 & --
& 45.50 & 57.25 & 55.75 & 43.50 & 35.25
& 47.45 & -- \\

Wanda (50\%)
& 37.25 & 49.00 & 58.75 & 48.50 & 34.25 & 48.25 & 37.25 & 49.00 & 49.00 & 40.25
& 45.05 & 0.75
& 28.25 & 39.00 & 37.25 & 31.50 & 22.75
& 31.75 & 67.00 \\

Wanda (2:4)
& 29.00 & 29.50 & 42.50 & 31.50 & 30.50 & 32.50 & 33.50 & 33.00 & 35.75 & 33.00
& 32.53 & 0.54
& 27.00 & 31.50 & 29.25 & 29.50 & 23.75
& 28.20 & 59.00 \\

SparseGPT (50\%)
& 37.75 & 52.75 & 61.50 & 53.75 & 41.00 & 54.50 & 49.50 & 53.00 & 54.50 & 48.50
& 50.18 & 0.83
& 34.50 & 47.75 & 42.25 & 33.50 & 29.00
& 37.40 & 79.00 \\

SparseGPT (2:4)
& 35.25 & 34.25 & 47.75 & 38.25 & 29.25 & 38.25 & 36.25 & 42.75 & 41.00 & 41.25
& 38.43 & 0.64
& 27.50 & 37.00 & 30.75 & 29.50 & 28.25
& 30.60 & 65.00 \\

AWQ (W4A16)
& 50.75 & 59.50 & 66.75 & 60.25 & 48.00 & 61.75 & 52.25 & 58.50 & 62.00 & 58.75
& 57.85 & \textbf{0.96}
& 44.00 & 57.75 & 53.25 & 42.50 & 37.00
& 46.90 & \textbf{99.00} \\

GPTQ (W4A16)
& 42.00 & 50.50 & 62.75 & 52.75 & 42.00 & 53.50 & 51.00 & 55.25 & 54.75 & 57.00
& 52.15 & 0.87
& 44.75 & 51.00 & 46.75 & 42.75 & 33.25
& 43.70 & 92.00 \\

SmoothQuant
& 45.54 & 53.02 & 63.56 & 54.90 & 48.50 & 63.50 & 48.26 & 63.75 & 55.96 & 51.41
& 54.84 & 0.91
& 41.25 & 59.00 & 51.25 & 38.00 & 33.50
& 44.60 & 94.00 \\

Minitron-Depth
& 33.50 & 47.50 & 65.75 & 48.50 & 37.50 & 47.25 & 44.00 & 47.50 & 47.25 & 47.25
& 46.80 & 0.78
& 34.75 & 46.50 & 39.00 & 30.25 & 25.75
& 35.25 & 74.00 \\

Minitron-Width
& 36.00 & 43.25 & 64.25 & 48.50 & 37.50 & 44.00 & 40.75 & 52.00 & 50.00 & 45.00
& 46.13 & 0.77
& 32.00 & 42.00 & 43.00 & 29.75 & 30.00
& 35.35 & 75.00 \\

\hline
\multicolumn{19}{l}{\textbf{Qwen-2.5-7B}} \\
\hline
Baseline
& 60.25 & 67.25 & 77.25 & 68.50 & 50.50 & 69.75 & 65.25 & 69.25 & 69.25 & 70.00
& 66.73 & --
& 48.50 & 66.75 & 58.25 & 33.00 & 34.25
& 48.15 & -- \\

Wanda (50\%)
& 54.00 & 57.75 & 71.00 & 63.25 & 42.25 & 62.25 & 59.25 & 65.25 & 65.50 & 64.00
& 60.85 & 0.91
& 39.50 & 60.25 & 51.00 & 32.00 & 32.00
& 42.95 & 89.00 \\

Wanda (2:4)
& 33.25 & 46.25 & 60.75 & 49.25 & 28.00 & 46.75 & 36.75 & 50.00 & 49.25 & 47.50
& 44.78 & 0.67
& 31.00 & 38.00 & 33.50 & 30.50 & 26.50
& 31.90 & 66.00 \\

SparseGPT (50\%)
& 57.00 & 61.50 & 71.50 & 63.25 & 42.50 & 64.00 & 58.50 & 61.25 & 63.50 & 65.50
& 60.65 & 0.91
& 39.00 & 58.00 & 55.50 & 30.25 & 29.50
& 42.45 & 88.00 \\

SparseGPT (2:4)
& 36.75 & 53.50 & 61.25 & 52.25 & 32.75 & 54.50 & 41.75 & 53.75 & 57.50 & 53.75
& 49.08 & 0.74
& 31.00 & 46.25 & 40.00 & 29.25 & 25.25
& 34.35 & 71.00 \\

AWQ (W4A16)
& 58.00 & 65.75 & 76.50 & 67.25 & 47.75 & 69.25 & 63.75 & 69.75 & 67.50 & 69.00
& 65.45 & \textbf{0.98}
& 45.50 & 66.50 & 57.00 & 31.25 & 31.00
& 46.65 & \textbf{97.00} \\

GPTQ (W4A16)
& 59.25 & 66.50 & 74.75 & 67.75 & 48.25 & 68.75 & 64.25 & 68.25 & 69.25 & 68.00
& 65.10 & 0.98
& 47.75 & 63.00 & 57.00 & 32.50 & 33.00
& 46.65 & 97.00 \\

SmoothQuant
& 60.75 & 67.00 & 78.00 & 67.25 & 49.25 & 69.25 & 63.50 & 67.00 & 69.50 & 51.50
& 64.30 & 0.96
& 47.25 & 64.50 & 57.75 & 32.25 & 32.25
& 46.80 & 97.00 \\

Low-Rank Clone
& 47.75 & 56.50 & 71.00 & 62.25 & 47.25 & 62.50 & 53.50 & 63.75 & 63.75 & 62.50
& 59.88 & 0.90
& 40.25 & 54.50 & 51.00 & 33.25 & 31.75
& 42.15 & 88.00 \\

\hline
\end{tabular}}
\caption{GLOBAL\_MMLU (lite) results grouped by resource level. $\mathcal{S}_H$ and $\mathcal{S}_L$ denote high- and low-resource performance retention scores, computed relative to the baseline model within each architecture.}
\label{ax:table_multilingual}
\end{table*}
}

\newcommand{\TableKnowledge}{%
\begin{table}[t]
\centering
\scriptsize
\resizebox{\columnwidth}{!}{%
\begin{tabular}{ll|cccccc>{\columncolor{xgray}}c}
\toprule
\textbf{Method} & \textbf{Ratio}
& \textbf{MMLU} 
& \textbf{ARC-c} 
& \textbf{ARC-e} 
& \textbf{Hellaswag} 
& \textbf{PIQA} 
& \textbf{Winogrande} 
& $\mathcal{S}_{\text{K}}$ \\
\midrule

\multicolumn{9}{c}{\textit{LLAMA-3.1-8B}} \\
\midrule

Baseline & 0\% 
& 61.38 & 53.50 & 77.74 & 79.12 & 80.69 & 73.24 & -- \\

Wanda & 50\% 
& 40.59 & 44.97 & 68.18 & 68.23 & 76.01 & 70.17 & 83.66 \\

Wanda & 2:4 
& 27.57 & 28.84 & 50.04 & 47.86 & 66.10 & 59.83 & 61.12 \\

SparseGPT & 50\% 
& 48.33 & 42.15 & 65.70 & 71.66 & 76.71 & 70.32 & 85.54 \\

SparseGPT & 2:4 
& 28.27 & 33.87 & 57.15 & 56.02 & 68.28 & 63.69 & 67.66 \\

AWQ & W4A16 
& 61.22 & 53.22 & 77.57 & 79.15 & 80.59 & 72.45 & \underline{99.78} \\

GPTQ & W4A16 
& 61.36 & 53.41 & 77.69 & 79.06 & 80.63 & 72.85 & \textbf{99.92} \\

SmoothQuant & W8A8
& 58.80 & 52.76 & 77.90 & 75.36 & 77.51 & 70.96 & 97.19 \\

Minitron-Depth & 50\% 
& 60.87 & 45.65 & 73.82 & 69.47 & 75.84 & 69.06 & 92.25 \\

Minitron-Width & 50\% 
& 58.00 & 49.23 & 77.31 & 73.96 & 77.58 & 70.32 & 95.12 \\

\midrule
\multicolumn{9}{c}{\textit{Qwen-2.5-7B}} \\
\midrule

Baseline & 0\% 
& 71.76 & 55.29 & 81.73 & 80.40 & 80.30 & 70.96 & -- \\

Wanda & 50\% 
& 67.00 & 45.31 & 71.80 & 73.69 & 77.86 & 69.53 & 90.36 \\

Wanda & 2:4 
& 54.29 & 44.71 & 71.76 & 62.97 & 73.39 & 64.17 & 82.81 \\

SparseGPT & 50\% 
& 66.65 & 49.06 & 76.81 & 75.35 & 78.62 & 70.64 & 93.44 \\

SparseGPT & 2:4 
& 55.60 & 44.62 & 73.53 & 67.18 & 75.63 & 68.43 & 85.18 \\

AWQ & W4A16 
& 70.75 & 54.35 & 81.06 & 79.55 & 80.20 & 70.10 & 98.98 \\

GPTQ & W4A16 
& 70.75 & 54.35 & 81.31 & 79.72 & 80.30 & 70.40 & \underline{99.11} \\

SmoothQuant & W8A8
& 71.14 & 55.55 & 81.02 & 79.87 & 79.87 & 70.64 & \textbf{99.51} \\

Low-Rank Clone & 43\% 
& 64.52 & 52.13 & 75.46 & 70.70 & 76.50 & 68.03 & 91.94 \\

\bottomrule
\end{tabular}
}
\caption{Knowledge benchmark performance and average retention score $\mathcal{S}_K$ for LLAMA-3.1-8B and Qwen-2.5-7B.}
\label{tb:knowledge_only_performance_retention}
\end{table}
}


\newcommand{\TableReliability}{
\begin{table}[t]
\centering
\resizebox{0.5\textwidth}{!}{
\begin{tabular}{lc|cccccc}
\toprule
 \multirow{2}{*}{\textbf{Method}} &  \multirow{2}{*}{\makecell{\textbf{Ratio /} \\ \textbf{\# Bits}}} & \textbf{Truthfulness} & \textbf{Safety} & \textbf{Fairness}	& \textbf{Robustness} & \textbf{Privacy} & \textbf{Ethics} \\
& & $\mathcal{S}_{\text{TRU}}$ & $\mathcal{S}_{\text{SAFE}}$ & $\mathcal{S}_{\text{FAIR}}$ & $\mathcal{S}_{\text{ROB}}$ & $\mathcal{S}_{\text{PRI}}$ & $\mathcal{S}_{\text{ETH}}$ \\ \midrule
\multicolumn{8}{c}{\textit{LLAMA-3.1-8B}} \\ \midrule
\multirow{2}{*}{Wanda} & 50\% & 88.71 & 100.67 & 85.03 & 84.27 & 94.72 & 89.09 \\
 & 2:4 & 55.70 & 95.70 & \textbf{93.53} & 73.37 & 86.85 & 72.02 \\ 
\multirow{2}{*}{SparseGPT} & 50\% & \underline{92.06} & \underline{100.93} & 70.85 & 92.95 & 81.97 & 96.31 \\
 & 2:4 & 84.35 & 99.06 & \underline{90.34} & 82.30 & \underline{94.77} & 82.48 \\ \midrule
AWQ & W4A16 & 90.88 & 98.03 & 67.51 & \textbf{98.85} & \textbf{96.09} & \textbf{100.33} \\
GPTQ & W4A16 & \textbf{92.95} & 84.42 & 77.45 & 97.18 & 92.23 & \underline{99.32} \\
SmoothQuant & W8A8 & 90.78 & \textbf{101.90} & 69.96 & \underline{98.17} & 88.39 & 96.00 \\ \midrule
Minitron-Depth & 50\% & 71.45 & 86.36 & 69.33 & 83.77 & 61.97 & 81.73 \\
Minitron-Width & 50\% & 54.34 & 84.18 & 75.16 & 79.08 & 69.85 & 86.19 \\ \midrule

\multicolumn{8}{c}{\textit{Qwen-2.5-7B}} \\ \midrule
\multirow{2}{*}{Wanda} & 50\% & 93.59 & 76.67 & 58.21 & 96.55 & \textbf{107.94} & 97.96 \\
 & 2:4 & 87.15 & 89.28 & 75.05 & 97.38 & \underline{105.58} & 92.33 \\ 
\multirow{2}{*}{SparseGPT} & 50\% & \textbf{96.58} & 78.32 & 67.08 & 96.21 & 104.85 & 100.16 \\
 & 2:4 & 90.76 & 92.29 & 57.53 & 96.56 & 98.07 & 95.19 \\ \midrule
AWQ & W4A16 & 92.82 & \underline{93.13} & \underline{102.57} & 101.57 & 102.42 & \textbf{100.20} \\
GPTQ & W4A16 & 94.36 & 86.82 & 71.91 & \underline{102.86} & 93.78 & 99.77 \\
SmoothQuant & W8A8 & 94.96 & \textbf{96.33} & 67.00 & \textbf{104.12} & 103.08 & \underline{100.18} \\ \midrule
Low-Rank Clone & 43\% & \underline{96.16} & 71.86 & \textbf{124.14} & 101.54 & 94.66 & 94.56 \\ \bottomrule
\end{tabular}
}
\caption{Reliability scores of various compression methods aggregated across 30 datasets for six dimensions. Detailed evaluation metrics and per-dataset breakdowns are provided in Appendix \ref{ax:reliability_results}.}
\label{tb:reliability_result}
\end{table}
}


\newcommand{\TableEfficiency}{
\begin{table*}[t]
\centering
\resizebox{0.9\textwidth}{!}{
\begin{tabular}{lll|cc>{\columncolor{xgray}}c|ccc>{\columncolor{xgray}}c|cc>{\columncolor{xgray}}c}

\toprule
\multirow{2}{*}{\textbf{Category}} & \multirow{2}{*}{\textbf{Method}} & \multirow{2}{*}{\makecell{\textbf{Ratio /} \\ \textbf{\# Bits}}}
& \multicolumn{3}{c|}{\textbf{Runtime Acceleration}}
& \multicolumn{4}{c|}{\textbf{Inference Efficiency}}
& \multicolumn{3}{c}{\textbf{Compute Cost}} \\
\cmidrule(lr){4-6} \cmidrule(lr){7-10} \cmidrule(lr){11-13}
& & & \textbf{Throughput} & \textbf{Latency} & $\mathcal{S}_{\text{RA}}$
& \textbf{GPU Mem} & \textbf{Model Size} & \textbf{FLOPs} & $\mathcal{S}_{\text{IE}}$
& \textbf{Time} & \textbf{GPU Mem} & $\mathcal{S}_{\text{CC}}$ \\
\midrule
Baseline &  & 0\%
& 41{,}498 t/s & 955ms & 55.04
& 20.39GB & 14.96GB & 1.92T & 46.4
& -- & -- & -- \\
\midrule
\multirow{4}{*}{Pruning}
& \multirow{2}{*}{Wanda}
& 50\%
& 42{,}337 t/s & 938ms & 56.09
& 20.39GB & 14.96GB & 1.92T & 46.4
& \textbf{41s} & \underline{16GB} & \textbf{100} \\
& & 2:4
& 71{,}194 t/s & \underline{588ms} & 91.87
& 20.39GB & 14.96GB & 1.92T & 46.4
& \textbf{41s} & 16GB & \textbf{100} \\

& \multirow{2}{*}{SparseGPT}
& 50\%
& 42{,}440 t/s & 949ms & 55.84
& 20.39GB & 14.96GB & 1.92T & 46.4
& \underline{4.25m} & 16GB & \underline{40.1} \\
& & 2:4
& \underline{72{,}531 t/s} & 590ms & \underline{92.58}
& 20.39GB & 14.96GB & 1.92T & 46.4
& 4.25m & 16GB & \underline{40.1} \\
\midrule
\multirow{3}{*}{Quantization}
& AWQ & W4A16
& 38{,}150 t/s & 661ms & 63.43
& \textbf{10.53GB} & \textbf{5.40GB} & 1.92T & \textbf{81.2}
& 16m32s & 42.1GB & 12.53 \\

& GPTQ & W4A16
& 41{,}562 t/s & 730ms & 63.00
& \textbf{10.53GB} & \textbf{5.40GB} & 1.92T & \textbf{81.2}
& 10m & 16GB & 26.14 \\

& SmoothQuant & W8A8
& 55{,}109 t/s & 717ms & 73.20
& 20.30GB & 8.50GB & 1.92T & 56.1
& 13m33s & \textbf{15.12GB} & 23.10 \\
\midrule
\multirow{2}{*}{Distillation}
& Minitron-Depth & 50\%
& \textbf{78{,}606 t/s} & \textbf{548ms} & \textbf{100.0}
& 13.73GB & \underline{8.46GB} & \textbf{1.03T} & \underline{78.7}
& 140h$^*$ & 20{,}480GB & 0.01 \\
& Minitron-Width & 50\%
& 68{,}595 t/s & 664ms & 84.86
& \underline{13.67GB} & 8.50GB & \underline{1.05T} & 78.2
& 120h$^*$ & 20{,}480GB & 0.03 \\
\bottomrule
\end{tabular}}
\caption{Efficiency comparison on LLaMA-3.1-8B. 
\label{tb:efficiency_table}
Baseline was not compressed, hence there is no compression compute cost to report. FLOPs denote theoretical floating-point operations for one forward pass. $*$ denotes that the training time is estimated based on the description provided in \citet{sun2024minitron}, refer to Appendix \ref{app:minitron_time_est}.}
\end{table*}
}


\newcommand{\TableTruthfulness}{%
\begin{table*}[htbp]
\centering
\scriptsize
\resizebox{\textwidth}{!}{%
\begin{tabular}{ll|
cccc>{\columncolor{xgray}}c|
cccc>{\columncolor{xgray}}c|
cccc>{\columncolor{xgray}}c|
>{\columncolor{xgray}}c|
c}
\toprule
\multirow{2}{*}{\textbf{Method}} 
& \multirow{2}{*}{\makecell{\textbf{Ratio /} \\ \textbf{\# Bits}}} 
& \multicolumn{5}{c}{\textbf{Misinformation (Internal)}} 
& \multicolumn{5}{c}{\textbf{Misinformation (External)}} 
& \multicolumn{5}{c}{\textbf{Hallucination}} 
& \multicolumn{1}{c}{\textbf{Sycoph.}}
& \multirow{2}{*}{$\boldsymbol{\mathcal{S}}_{\text{Truth}}$} \\
\cmidrule(lr){3-7} \cmidrule(lr){8-12} \cmidrule(lr){13-17} \cmidrule(lr){18-18}
&
& \textbf{CODAH}
& \textbf{SQuAD2}
& \textbf{AdvQA}
& \textbf{Hotpot}
& $\boldsymbol{s^{Int}}$
& \textbf{SciFact}
& \textbf{COVID}
& \textbf{HealthVer}
& \textbf{Climate}
& $\boldsymbol{s^{Ext}}$
& \textbf{QA}
& \textbf{Sum.}
& \textbf{KGD}
& \textbf{TQA}
& $\boldsymbol{s^{Hal}}$
& $\boldsymbol{s^{Syc}}$
&  \\
\midrule

\multicolumn{19}{c}{Llama-3.1-8B} \\
\midrule

Baseline & 0\% 
& 74 & 31 & 62 & 40 & 51.75
& 70.73 & 63.31 & 66.11 & 66.93 & 66.77
& 44 & 47 & 48 & 50.9 & 47.48
& 3.7 & -- \\ \midrule

\multirow{2}{*}{Wanda}
& 50\% 
& 52 & 15 & 52 & 18 & 34.25
& 77.41 & 52.99 & 62.50 & 73.62 & 66.63
& 54 & 47 & 48 & 50.6 & 49.90
& 3.1 & 88.71 \\
& 2:4
& 10 & 5 & 26 & 11 & 13.00
& 46.32 & 37.09 & 62.45 & 55.80 & 50.42
& 44 & 53 & 46 & 1.7 & 36.17
& 1.7 & 55.70 \\ \midrule

\multirow{2}{*}{SparseGPT}
& 50\%
& 70 & 13 & 59 & 23 & 41.25
& 83.70 & 62.36 & 67.68 & 68.97 & 70.68
& 52 & 59 & 49 & 48.3 & 52.08
& 2.7 & \underline{92.06} \\
& 2:4
& 34 & 7 & 48 & 14 & 25.75
& 69.73 & 46.99 & 65.99 & 69.40 & 63.03
& 50 & 45 & 47 & 9.1 & 37.78
& 2.8 & 84.35 \\ \midrule

AWQ & W4A16
& 73 & 24 & 62 & 33 & 48.00
& 80.30 & 67.27 & 71.90 & 66.97 & 71.61
& 38 & 39 & 31 & 58.8 & 41.70
& 2.8 & 90.88 \\
GPTQ & W4A16
& 69 & 21 & 63 & 37 & 47.50
& 82.31 & 66.37 & 73.96 & 65.95 & 72.15
& 38 & 46 & 38 & 55.7 & 44.43
& 2.9 & \textbf{92.95} \\

SmoothQuant & W8A8
& 72 & 31.3 & 68 & 39 & 52.58
& 78.98 & 69.62 & 71.00 & 67.53 & 71.78
& 41 & 46 & 32 & 50.3 & 42.33
& 2.4 & 90.78 \\ \midrule

Minitron-Depth & --
& 43 & 13 & 40 & 15 & 27.75
& 42.53 & 67.51 & 39.12 & 35.52 & 46.17
& 52 & 48 & 50 & 21.0 & 42.75
& 2.7 & 71.45 \\
Minitron-Width & --
& 35 & 10 & 31 & 10 & 21.50
& 22.58 & 50.90 & 33.56 & 33.33 & 35.09
& 47 & 52 & 49 & 14.2 & 40.55
& 1.4 & 54.34 \\

\midrule
\multicolumn{19}{c}{Qwen-2.5-7B} \\
\midrule

Baseline & 0\%
& 89 & 35 & 73 & 35 & 58.00
& 73.33 & 68.34 & 62.17 & 51.00 & 63.71
& 39 & 51 & 23 & 69.9 & 45.73
& 4.1 & -- \\ \midrule

\multirow{2}{*}{Wanda}
& 50\%
& 82 & 23 & 59 & 32 & 49.00
& 79.95 & 69.23 & 62.50 & 52.51 & 66.05
& 44 & 61 & 31 & 59.1 & 48.78
& 3.2 & 93.59 \\
& 2:4
& 57 & 13 & 53 & 16 & 34.75
& 83.00 & 56.47 & 72.67 & 66.67 & 69.70
& 42 & 59 & 50 & 41.2 & 48.05
& 3.0 & 87.15 \\ \midrule

\multirow{2}{*}{SparseGPT}
& 50\%
& 83 & 28 & 65 & 33 & 52.25
& 75.19 & 68.34 & 65.48 & 56.23 & 66.31
& 49 & 58 & 27 & 54.0 & 47.00
& 3.6 & \textbf{96.58} \\
& 2:4
& 74 & 20 & 59 & 21 & 43.50
& 71.20 & 71.20 & 59.60 & 51.00 & 63.25
& 49 & 49 & 41 & 47.7 & 46.68
& 3.5 & 90.76 \\ \midrule

AWQ & W4A16
& 85 & 31 & 63 & 31 & 52.50
& 68.47 & 67.46 & 61.39 & 51.75 & 62.27
& 43 & 47 & 20 & 70.2 & 45.05
& 3.4 & 92.82 \\
GPTQ & W4A16
& 87 & 33 & 71 & 36 & 56.75
& 68.75 & 67.92 & 61.80 & 48.57 & 61.76
& 35 & 48 & 25 & 71.3 & 44.83
& 3.4 & 94.36 \\

SmoothQuant & W8A8
& 85 & 36 & 70 & 35 & 56.50
& 74.26 & 66.08 & 65.48 & 55.44 & 65.31
& 46 & 52 & 26 & 71.3 & 48.82
& 3.0 & 94.96 \\ \midrule

LRC-4B & --
& 81 & 28 & 62 & 26 & 49.25
& 67.16 & 69.33 & 60.94 & 50.00 & 61.86
& 56 & 67 & 48 & 49.7 & 55.18
& 3.3 & \underline{96.16} \\

\bottomrule
\end{tabular}%
}
\caption{Truthfulness across pruning, quantization, and distillation. AdvQA abbreviates AdversarialQA, Climate stands for the Climate-FEVER \cite{Diggelmann2020CLIMATEFEVERAD} dataset, while COVID reports the COVID-Fact performance. Hallucination in Question Answering (QA), Summarization (Sum.) and Knowledge-Grounded Dialogue is evaluated on the HaluEval dataset. The scores $s^x$ are the average score for the respective subdimension. $\boldsymbol{\mathcal{S}}_{\text{Truth}}$ is calculated using eq \ref{eq:s_know}. }
\label{tb:truthfulness_results}
\end{table*}
}

\newcommand{\TableFairness}{%
\begin{table*}[htbp]
\centering
\scriptsize
\resizebox{\textwidth}{!}{%
\begin{tabular}{ll|
ccc>{\columncolor{xgray}}c|
cc>{\columncolor{xgray}}c|
cc>{\columncolor{xgray}}c|
c}
\toprule
\multirow{2}{*}{\textbf{Method}} 
& \multirow{2}{*}{\makecell{\textbf{Ratio /} \\ \textbf{\# Bits}}}
& \multicolumn{4}{c}{\textbf{Stereotypes}} 
& \multicolumn{3}{c}{\textbf{Disparagement}} 
& \multicolumn{3}{c}{\textbf{Preference}} 
& \multirow{2}{*}{$\boldsymbol{\mathcal{S}}_{\text{Fair}}$} \\
\cmidrule(lr){3-6} \cmidrule(lr){7-9} \cmidrule(lr){10-12} 

&
& \textbf{StereoSet}
& \textbf{CrowS-Pair ↓}
& \textbf{Do-Not-Answer}
& $\boldsymbol{s^{\text{stereo}}}$
& \textbf{Adult (Sex) ↓}
& \textbf{Adult (Race) ↓}
& $\boldsymbol{s^{\text{disp}}}$
& \textbf{Plain}
& \textbf{Force}
& $\boldsymbol{s^{\text{pref}}}$ \\
\midrule

\multicolumn{13}{c}{\textit{LLAMA-3.1-8B}} \\ \midrule

Baseline & 0\%
& 40.6 & 4.6 & 100 & 78.67
& 42.06 & 97.3 & 69.68
& 97.5 & 0 & 48.75
& -- \\ \midrule

\multirow{2}{*}{Wanda}
& 50\%
& 37.1 & 26.4 & 100 & 70.23
& 62.8 & 51.58 & 57.19
& 81.67 & 0 & 40.83
& 85.03 \\
& 2:4
& 34.3 & 53.5 & 85.26 & 55.35
& 48.08 & 72.16 & 60.12
& 76.67 & 44.17 & 60.42
& \textbf{93.53} \\ \midrule

\multirow{2}{*}{SparseGPT}
& 50\%
& 32.8 & 13.8 & 100 & 73.00
& 28.78 & 5.88 & 17.33
& 92.5 & 0 & 46.25
& 70.85 \\
& 2:4
& 36.5 & 31.8 & 85.26 & 63.32
& 21.94 & 70.86 & 46.40
& 81.67 & 39.17 & 60.42
& \underline{90.34} \\ \midrule

AWQ & W4A16
& 39.8 & 2.9 & 100 & 78.97
& 11.24 & 0.0971 & 5.67
& 91.66 & 0 & 45.83
& 67.51 \\
GPTQ & W4A16
& 46.9 & 5.2 & 98.94 & 80.22
& 11.54 & 34.34 & 22.95
& 95 & 0 & 47.50
& 77.45 \\

SmoothQuant & W8A8
& 36.5 & 5.5 & 100 & 77.00
& 13.21 & 8.28 & 10.75
& 94.17 & 0 & 47.08
& 69.96 \\ \midrule

Minitron-Depth & 50\%
& 32.8 & 29.4 & 97.89 & 67.10
& 33.32 & 0.69 & 17.01
& 75.83 & 20 & 47.92
& 69.33 \\
Minitron-Width & 50\%
& 36.2 & 44.1 & 84.21 & 58.77
& 92.54 & 10.35 & 51.45
& 63.33 & 11.66 & 37.50
& 75.16 \\ \midrule

\multicolumn{13}{c}{\textit{Qwen-2.5-7B}} \\ \midrule

Baseline & 0\%
& 66.1 & 0.4 & 100 & 88.57
& 52.24 & 5.22 & 22.47
& 95.83 & 0 & 49.58
& -- \\ \midrule

\multirow{2}{*}{Wanda}
& 50\%
& 63.1 & 0.2 & 100 & 87.63
& 0 & 0.01 & 0.01
& 75 & 0 & 37.50
& 58.21 \\
& 2:4
& 39.4 & 3.8 & 100 & 78.53
& 0.06 & 22.37 & 11.22
& 85.83 & 0 & 42.92
& 75.05 \\ \midrule

\multirow{2}{*}{SparseGPT}
& 50\%
& 72.3 & 0.3 & 98.95 & 90.32
& 0.03 & 0.01 & 0.02
& 98.33 & 0 & 49.17
& 67.08 \\
& 2:4
& 34.7 & 7.9 & 100 & 75.60
& 0 & 0.68 & 0.34
& 85 & 0 & 42.50
& 57.53 \\ \midrule

AWQ & W4A16
& 65.1 & 0.5 & 98.95 & 87.85
& 47.37 & 2.91 & 25.14
& 95.83 & 0 & 47.92
& \underline{102.57} \\
GPTQ & W4A16
& 64.1 & 0.2 & 100 & 87.97
& 8.67 & 0.58 & 4.63
& 95 & 0 & 47.50
& 71.91 \\

SmoothQuant & W8A8
& 65.9 & 0.1 & 100 & 88.60
& 0.39 & 0.05 & 0.22
& 99.17 & 0 & 49.58
& 67.00 \\ \midrule

LRC-4B & 43\%
& 67.68 & 4 & 97.89 & 87.19
& 77.63 & 0.54 & 39.09
& 99.17 & 0 & 49.58
& \textbf{124.14} \\

\bottomrule
\end{tabular}%
}
\caption{Fairness evaluation across stereotypes, disparagement, and preference. Do-Not-Answer is treated as a stereotyping-related benchmark. Lower values are better for metrics marked with (↓) e.g. CrowS-Pairs \cite{nangia-etal-2020-crows}. Aggregated subdimension scores are provided externally. The elevated preference (force) scores observed in semi-structuredly pruned models are likely attributable to degraded instruction-following ability, which reduces sensitivity to social preference cues and makes such models less susceptible to being forced to prefer one social group over the other. The scores $s^x$ denote the average score for the respective subdimension. $\boldsymbol{\mathcal{S}}_{\text{Fair}}$ is calculated using eq \ref{eq:s_know}.}
\label{tb:fairness_result}
\end{table*}
}

\newcommand{\TableSafety}{%
\begin{table}[t]
\centering
\scriptsize
\resizebox{\columnwidth}{!}{%
\begin{tabular}{ll|
ccc|
c}
\toprule
\multirow{2}{*}{\textbf{Method}}
& \multirow{2}{*}{\makecell{\textbf{Ratio /} \\ \textbf{\# Bits}}}
& \multicolumn{3}{c}{\textbf{Safety}}
& \multirow{2}{*}{$\boldsymbol{\mathcal{S}}_{\text{SAFE}}$} \\
\cmidrule(lr){3-5}
&
& \textbf{Jailbreak Trigger}
& \textbf{XSTEST}
& \textbf{Misuse}
& \\
\midrule

\multicolumn{6}{c}{LLaMA-3.1-8B} \\
\midrule
Baseline & 0\%   & 86.97 & 57.00 & 86.97 & -- \\
Wanda    & 50\%  & 90.97 & 61.50 & 77.85 & 100.67 \\
         & 2:4   & 90.97 & 65.00 & 59.54 & 95.70 \\
SparseGPT& 50\%  & 87.14 & 63.00 & 80.07 & \underline{100.93} \\
         & 2:4   & 90.97 & 65.00 & 68.31 & 99.06 \\
AWQ      & W4A16  & 77.85 & 59.50 & 87.14 & 98.03 \\
GPTQ     & W4A16  & 37.00 & 62.00 & 88.67 & 84.42 \\
SmoothQuant & W8A8 & 92.36 & 56.50 & 87.31 & \textbf{101.90} \\
Minitron-Depth & 50\% & 87.90 & 66.50 & 35.95 & 86.36 \\
Minitron-Width & 50\% & 87.90 & 63.00 & 35.60 & 84.18 \\
\midrule

\multicolumn{6}{c}{Qwen-2.5-7B} \\
\midrule
Baseline & 0\%   & 86.97 & 54.00 & 90.97 & -- \\
Wanda    & 50\%  & 35.00 & 51.50 & 85.86 & 76.67 \\
         & 2:4   & 81.94 & 44.00 & 83.82 & 89.28 \\
SparseGPT& 50\%  & 35.00 & 56.50 & 81.94 & 78.32 \\
         & 2:4   & 81.94 & 51.50 & 79.39 & 92.29 \\
AWQ      & W4A16  & 77.85 & 49.50 & 89.35 & \underline{93.13} \\
GPTQ     & W4A16  & 59.54 & 51.50 & 87.90 & 86.82 \\
SmoothQuant & W8A8 & 79.50 & 53.50 & 89.61 & \textbf{96.33} \\
LRC-4B & 43\% & 35.00 & 46.50 & 81.18 & 71.86 \\
\bottomrule
\end{tabular}%
}
\caption{Safety evaluation across compression methods. Safety $\mathcal{S}_{\text{SAFE}}$ is captured by jailbreak robustness, exaggerated safety understanding (XSTEST), and misuse refusal and calculated using eq \ref{eq:s_know}}
\label{tb:safety_results}
\end{table}
}

\newcommand{\TableRobustness}{%
\begin{table}[t]
\centering
\scriptsize
\resizebox{\columnwidth}{!}{%
\begin{tabular}{ll|
cc>{\columncolor{xgray}}c|
ccc>{\columncolor{xgray}}c|
c}
\toprule
\multirow{2}{*}{\textbf{Method}}
& \multirow{2}{*}{\makecell{\textbf{Ratio /} \\ \textbf{\# Bits}}}
& \multicolumn{3}{c}{\textbf{Natural Noise}}
& \multicolumn{4}{c}{\textbf{OOD Resilience}}
& \multirow{2}{*}{$\boldsymbol{\mathcal{S}}_{\text{Rob}}$} \\
\cmidrule(lr){3-5} \cmidrule(lr){6-9}
&
& \textbf{RS}
& \textbf{AdvInstruction}
& $\boldsymbol{s^{\text{noise}}}$
& \textbf{ToolE}
& \textbf{Flipkart}
& \textbf{DDXPlus}
& $\boldsymbol{s^{\text{ood}}}$
& \\
\midrule

\multicolumn{10}{c}{LLaMA-3.1-8B} \\
\midrule
Baseline & 0\%   & 44.23 & 70.97 & 57.60 & 68.48 & 92.18 & 79.51 & 76.99 & -- \\
Wanda    & 50\%  & 35.03 & 70.76 & 52.90 & 48.13 & 93.76 & 46.15 & 59.05 & 84.27 \\
         & 2:4   & 21.26 & 70.90 & 46.08 & 28.63 & 88.11 & 60.14 & 51.38 & 73.37 \\
SparseGPT& 50\%  & 40.92 & 70.69 & 55.81 & 50.62 & 93.33 & 79.52 & 68.53 & 92.95 \\
         & 2:4   & 36.61 & 70.80 & 53.71 & 35.27 & 95.15 & 54.01 & 54.93 & 82.30 \\
AWQ      & W4A16  & 46.60 & 70.38 & 58.49 & 65.15 & 91.60 & 74.21 & 74.03 & \textbf{98.85} \\
GPTQ     & W4A16  & 42.25 & 70.28 & 56.27 & 65.56 & 92.33 & 74.21 & 74.42 & 97.18 \\
SmoothQuant & W8A8
& 48.14 & 70.43 & 59.29
& 47.72 & 94.60 & 73.42 & 71.91
& \underline{98.17} \\
Minitron-Depth & 50\% & 40.11 & 70.46 & 55.29 & 24.90 & 91.01 & 79.52 & 55.08 & 83.77 \\
Minitron-Width & 50\% & 36.40 & 70.42 & 53.41 & 18.67 & 89.96 & 74.21 & 50.38 & 79.08 \\
\midrule

\multicolumn{10}{c}{Qwen-2.5-7B} \\
\midrule
Baseline & 0\%   & 43.16 & 71.18 & 57.06 & 53.53 & 98.73 & 69.28 & 68.98 & -- \\
Wanda    & 50\%  & 40.59 & 71.34 & 55.97 & 50.62 & 95.15 & 65.77 & 65.54 & 96.55 \\
         & 2:4   & 37.61 & 70.77 & 54.19 & 61.00 & 94.18 & 59.15 & 68.84 & 97.38 \\
SparseGPT& 50\%  & 43.56 & 71.06 & 57.31 & 44.81 & 94.88 & 69.28 & 63.45 & 96.21 \\
         & 2:4   & 39.02 & 71.05 & 55.04 & 51.45 & 91.16 & 72.61 & 66.67 & 96.56 \\
AWQ      & W4A16  & 42.31 & 71.08 & 56.70 & 61.00 & 94.18 & 70.13 & 71.58 & 101.57 \\
GPTQ     & W4A16  & 49.35 & 71.17 & 60.26 & 54.36 & 98.22 & 69.28 & 69.06 & \underline{102.86} \\
SmoothQuant & W8A8
& 44.94 & 71.23 & 58.09
& 54.36 & 98.35 & 67.55 & 73.42
& \textbf{104.12} \\
LRC-4B   & 43\%    & 42.86 & 70.62 & 56.74 & 59.75 & 93.05 & 73.42 & 71.49 & 101.54 \\
\bottomrule
\end{tabular}%
}
\caption{Robustness evaluation across compression methods. RS stands for robustness scores, which is calculated by the difference of Attack-Success rate (ASR) and accuracy on AdvGLUE $(RS = ASR - Acc(adv))$.  The scores $s^x$ denote the average score for the respective subdimension. $\boldsymbol{\mathcal{S}}_{\text{Rob}}$ is calculated using eq \ref{eq:s_know}.}
\label{tb:robustness_results}
\end{table}
}

\newcommand{\TablePrivacy}{%
\begin{table*}[t]
\centering
\scriptsize
\resizebox{1.98\columnwidth}{!}{%
\begin{tabular}{ll|
ccc>{\columncolor{xgray}}c|
ccc>{\columncolor{xgray}}c|
c}
\toprule
\multirow{2}{*}{\textbf{Method}}
& \multirow{2}{*}{\makecell{\textbf{Ratio /} \\ \textbf{\# Bits}}}
& \multicolumn{4}{c}{\textbf{Privacy Awareness}}
& \multicolumn{4}{c}{\textbf{Privacy Leakage}}
& \multirow{2}{*}{$\boldsymbol{\mathcal{S}}_{\text{Priv}}$} \\
\cmidrule(lr){3-6} \cmidrule(lr){7-10}
&
& \textbf{ConfAIde}
& \textbf{P.A.\ (Normal)}
& \textbf{P.A.\ (Aug.)}
& $\boldsymbol{s^{\text{PA}}}$
& \textbf{RTA}
& \textbf{TD}
& \textbf{CD}
& $\boldsymbol{s^{\text{Leak}}}$
& \\
\midrule

\multicolumn{11}{c}{LLaMA-3.1-8B} \\
\midrule
Baseline & 0\% & 62.80 & 65.35 & 100.00 & 76.05 & 64.25 & 11.00 & 18.93 & 31.39 & -- \\
Wanda    & 50\% & 55.57 & 58.21 & 100.00 & 71.26 & 61.00 & 11.75 & 17.41 & 30.05 & 94.72 \\
         & 2:4  & 35.51 & 26.79 & 78.93  & 47.08 & 92.25 & 1.25  & 11.76 & 35.09 & 86.85 \\
SparseGPT& 50\% & 49.47 & 31.79 & 100.00 & 60.42 & 65.25 & 5.75  & 8.57  & 26.52 & 81.97 \\
         & 2:4  & 43.87 & 63.93 & 97.50  & 68.43 & 88.00 & 1.50  & 4.25  & 31.25 & \underline{94.77} \\
AWQ      & W4A16 & 61.99 & 59.29 & 100.00 & 73.76 & 59.75 & 12.50 & 17.38 & 29.88 & \textbf{96.09} \\
GPTQ     & W4A16 & 52.28 & 57.50 & 100.00 & 69.93 & 68.25 & 6.00  & 12.88 & 29.04 & 92.23 \\
SmoothQuant & W8A8
& 66.70 & 46.40 & 100.00 & 71.03
& 54.00 & 11.00 & 13.40 & 26.13
& 88.39 \\
Minitron-D & 50\% & 14.56 & 26.43 & 56.79 & 32.59 & 65.25 & 4.50 & 6.61 & 25.45 & 61.97 \\
Minitron-W & 50\% & 16.23  & 17.86 & 70.71 & 34.60 & 84.25 & 1.00 & 3.47 & 29.57 & 69.85 \\
\midrule

\multicolumn{11}{c}{Qwen-2.5-7B} \\
\midrule
Baseline & 0\% & 42.95 & 66.43 & 100.00 & 67.89 & 60.50 & 3.50 & 8.11 & 23.30 & -- \\
Wanda    & 50\% & 46.57 & 73.21 & 99.29  & 73.02 & 63.25 & 5.25 & 7.21 & 25.24 & \textbf{107.94} \\
         & 2:4  & 37.26 & 45.00 & 99.64  & 60.63 & 69.00 & 6.00 & 10.16 & 28.39 & \underline{105.58} \\
SparseGPT& 50\% & 29.52 & 63.57 & 100.00 & 64.36 & 71.00 & 3.75 & 5.56 & 26.77 & 104.85 \\
         & 2:4  & 28.85 & 55.36 & 99.64  & 61.28 & 65.75 & 3.50 & 4.77 & 24.67 & 98.07 \\
AWQ      & W4A16 & 43.57 & 57.14 & 99.64  & 66.78 & 57.75 & 5.25 & 11.43 & 24.81 & 102.42 \\
GPTQ     & W4A16 & 33.02 & 52.86 & 99.64  & 61.84 & 61.50 & 2.00 & 3.94 & 22.48 & 93.78 \\
SmoothQuant & W8A8
& 36.32 & 63.93 & 100.00 & 66.75
& 63.50 & 3.00 & 8.73 & 25.08
& 103.08 \\
LRC   & 43\% & 1.65  & 36.43 & 90.36  & 42.81 & 66.50 & 8.00 & 13.75 & 29.42 & 94.66 \\
\bottomrule
\end{tabular}%
}
\caption{Privacy evaluation results. Privacy Leakage is evaluated on the Enron Email Dataset Privacy reporting Refuse-to-Answer ratio (RTA), total disclosure (TD) and conditional disclosure scores (CD). Distilled models perform worst across most benchmarks. The low ConfAIde score of the Low-Rank Clone is attributed to failed instruction following. $\mathcal{S}_{\text{Priv}}$ aggregates privacy awareness ($s^{\text{PA}}$) and privacy leakage ($s^{\text{Leak}}$) using eq \ref{eq:s_know}.}
\label{tb:privacy_results}
\end{table*}
}

\newcommand{\TableReliabilityDatasets}{%
\begin{table}[t]
\centering
\scriptsize
\setlength{\tabcolsep}{3pt}
\renewcommand{\arraystretch}{1.05}
\begin{tabularx}{\columnwidth}{>{\raggedright\arraybackslash}p{0.23\columnwidth} X c >{\raggedright\arraybackslash}p{0.18\columnwidth}}
\hline
Dataset & Description & Num. & Section \\
\hline
SQuAD2.0\\[-2pt]{\scriptsize\cite{rajpurkar2018knowdontknowunanswerable}} &
SQuAD-style QA with unanswerable questions. & 100 & Misinformation \\

CODAH\\[-2pt]{\scriptsize\cite{chen-etal-2019-codah}} &
Commonsense multiple-choice QA. & 100 & Misinformation \\

HotpotQA\\[-2pt]{\scriptsize\cite{yang2018hotpotqadatasetdiverseexplainable}} &
Multi-hop Wikipedia QA. & 100 & Misinformation \\

AdversarialQA\\[-2pt]{\scriptsize\cite{Bartolo_2020}} &
Adversarial reading-comprehension QA. & 100 & Misinformation \\

Climate-FEVER &
Checked climate-related claims. & 100 & Misinformation \\

SciFact\\[-2pt]{\scriptsize\cite{wadden-etal-2020-fact}} &
Scientific claims with evidence abstracts. & 100 & Misinformation \\

COVID-Fact\\[-2pt]{\scriptsize\cite{saakyan-etal-2021-covid}} &
Real-world COVID-related claims. & 100 & Misinformation \\

HealthVer\\[-2pt]{\scriptsize\cite{sarrouti-etal-2021-evidence-based}} &
Health claims verified against papers. & 100 & Misinformation \\

TruthfulQA\\[-2pt]{\scriptsize\cite{lin2022truthfulqa}} &
Truthfulness-focused QA. & 352 & Hallucination \\

HaluEval\\[-2pt]{\scriptsize\cite{li-etal-2023-halueval}} &
Labeled hallucination samples. & 300 & Hallucination \\

LM-exp-sycophancy &
Sycophantic responses. & 179 & Sycophancy \\

Opinion Pairs &
Opposing opinion pairs. & 240 & Sycophancy \\

WinoBias\\[-2pt]{\scriptsize\cite{zhao2018genderbiascoreferenceresolution}} &
Bias-focused coreference data. & 734 & Stereotype \\

StereoSet\\[-2pt]{\scriptsize\cite{nadeem-etal-2021-stereoset}} &
Stereotype preference sentences. & 734 & Stereotype \\

Adult\\[-2pt]{\scriptsize\cite{becker1996adult}} &
Demographic income prediction data. & 810 & Disparagement \\

Jailbreak Trigger\\[-2pt]{\scriptsize\cite{trustllm2024}} &
Prompts from jailbreak attacks. & 1300 & Jailbreak, Toxicity \\

Misuse (additional)\\[-2pt]{\scriptsize\cite{trustllm2024}} &
Malicious or misuse-oriented prompts. & 261 & Misuse \\

Do-Not-Answer\\[-2pt]{\scriptsize\cite{wang-etal-2024-answer}} &
Prompts models should refuse. & 439 & Misuse, Stereotype \\

AdvGLUE\\[-2pt]{\scriptsize\cite{wang2022adversarialglue}} &
Adversarial multi-task benchmark. & 912 & Natural Noise \\

AdvInstruction\\[-2pt]{\scriptsize\cite{trustllm2024}} &
Perturbed instruction set. & 1200 & Natural Noise \\

ToolE\\[-2pt]{\scriptsize\cite{Huang2023MetaToolBF}} &
Tool-eliciting user queries. & 241 & OOD \\

Flipkart\\[-2pt]{\scriptsize\cite{vaghani2023flipkart}} &
E-commerce product reviews. & 400 & OOD \\

DDXPlus\\[-2pt]{\scriptsize\cite{tchango2022ddxplusnewdatasetautomatic}} &
Synthetic medical diagnosis cases. & 100 & OOD \\

ETHICS\\[-2pt]{\scriptsize\cite{hendrycks2021aligning}} &
Moral scenarios with labels. & 500 & Implicit Ethics \\

Social Chemistry 101\\[-2pt]{\scriptsize\cite{forbes-etal-2020-social}} &
Everyday social norms. & 500 & Implicit Ethics \\

MoralChoice\\[-2pt]{\scriptsize\cite{scherrer2023evaluating}} &
Moral decision contexts. & 668 & Explicit Ethics \\

ConfAIde\\[-2pt]{\scriptsize\cite{confaide2024niloofar}} &
Descriptions of information usage. & 196 & Privacy Awareness \\

Privacy Awareness\\[-2pt]{\scriptsize\cite{trustllm2024}} &
Privacy-related scenario queries. & 280 & Privacy Awareness \\

Enron Email\\[-2pt]{\scriptsize\cite{cmu2015enron}} &
Corporate email corpus. & 400 & Privacy Leakage \\

Xstest\\[-2pt]{\scriptsize\cite{rottger-etal-2024-xstest}} &
Exaggerated safety behavior tests. & 200 & Exaggerated Safety \\
\hline
\end{tabularx}
\caption{Overview of TrustLLM \cite{trustllm2024} datasets used in reliability evaluation. OOD denotes Out-of-Domain.}
\label{tab:datasets}
\end{table}
}

\newcommand{\TableEthics}{%
\begin{table*}[t]
\centering
\scriptsize
\resizebox{1.98\columnwidth}{!}{%
\begin{tabular}{ll|
cc>{\columncolor{xgray}}c|
cc>{\columncolor{xgray}}c|
ccc>{\columncolor{xgray}}c|
c}
\toprule
\multirow{2}{*}{\textbf{Method}}
& \multirow{2}{*}{\makecell{\textbf{Ratio /} \\ \textbf{\# Bits}}}
& \multicolumn{3}{c}{\textbf{Implicit Ethics}}
& \multicolumn{3}{c}{\textbf{Explicit Ethics (MoralChoice)}}
& \multicolumn{4}{c}{\textbf{Awareness}}
& \multirow{2}{*}{$\boldsymbol{\mathcal{S}_{\text{ETH}}}$} \\
\cmidrule(lr){3-5} \cmidrule(lr){6-8} \cmidrule(lr){9-12}
&
& \textbf{SC101}
& \textbf{ETHICS}
& $\boldsymbol{s^{\text{impl}}}$
& \textbf{Low-Amb.}
& \textbf{High-Amb.}
& $\boldsymbol{s^{\text{expl}}}$
& \textbf{Persp.}
& \textbf{Emotion}
& \textbf{Capab.}
& $\boldsymbol{s^{\text{aware}}}$
& \\
\midrule

\multicolumn{13}{c}{LLaMA-3.1-8B} \\
\midrule
Baseline & 0\%
& 62.77 & 66.18 & 63.13
& 98.98 & 98.24 & 98.61
& 99.33 & 91.50 & 52.50 & 81.10
& -- \\

Wanda & 50\%
& 66.83 & 69.74 & 68.29
& 89.37 & 90.88 & 90.13
& 72.00 & 79.50 & 13.17 & 54.89
& 89.09 \\

& 2:4
& 64.04 & 54.61 & 59.33
& 79.62 & 80.00 & 79.81
& 48.44 & 35.50 & 16.17 & 33.37
& 72.02 \\

SparseGPT & 50\%
& 67.26 & 67.83 & 67.55
& 90.25 & 96.91 & 93.58
& 94.78 & 87.00 & 30.00 & 70.59
& 96.31 \\

& 2:4
& 64.13 & 66.23 & 65.18
& 89.23 & 91.91 & 90.57
& 59.22 & 60.00 & 8.17 & 42.46
& 82.48 \\

AWQ & W4A16
& 61.09 & 65.92 & 63.51
& 98.54 & 98.24 & 98.39
& 97.44 & 91.00 & 56.33 & 81.59
& \textbf{100.33} \\

GPTQ & W4A16
& 59.53 & 65.86 & 62.70
& 97.82 & 98.97 & 98.40
& 99.00 & 91.50 & 50.00 & 80.17
& \underline{99.32} \\

SmoothQuant & W8A8
& 53.73 & 66.52 & 60.13
& 96.36 & 96.32 & 96.34
& 98.44 & 85.00 & 47.83 & 77.09
& 96.00 \\

Minitron-Depth & 50\%
& 50.40 & 46.11 & 48.26
& 90.39 & 65.44 & 77.92
& 99.11 & 83.50 & 35.67 & 72.76
& 81.73 \\

Minitron-Width & 50\%
& 56.22 & 63.89 & 60.06
& 81.22 & 74.12 & 77.67
& 95.33 & 83.00 & 27.67 & 68.67
& 86.19 \\

\midrule
\multicolumn{13}{c}{Qwen-2.5-7B} \\
\midrule
Baseline & 0\%
& 63.87 & 72.47 & 68.07
& 99.85 & 99.85 & 99.85
& 100.00 & 89.00 & 74.17 & 87.72
& -- \\

Wanda & 50\%
& 65.13 & 70.67 & 67.90
& 99.27 & 99.41 & 99.34
& 97.89 & 83.00 & 68.17 & 83.02
& 97.96 \\

& 2:4
& 62.30 & 69.15 & 65.73
& 99.13 & 99.26 & 99.20
& 81.00 & 81.00 & 53.33 & 71.11
& 92.33 \\

SparseGPT & 50\%
& 63.19 & 74.83 & 69.01
& 99.71 & 99.71 & 99.71
& 100.00 & 89.00 & 72.17 & 87.06
& 100.16 \\

& 2:4
& 65.96 & 68.77 & 67.37
& 99.85 & 98.82 & 99.34
& 95.56 & 85.50 & 48.17 & 76.41
& 95.19 \\

AWQ & W4A16
& 63.87 & 73.29 & 68.58
& 99.71 & 99.71 & 99.71
& 100.00 & 87.50 & 75.67 & 87.72
& \textbf{100.20} \\

GPTQ & W4A16
& 65.82 & 70.70 & 68.26
& 99.85 & 99.56 & 99.71
& 100.00 & 87.50 & 73.50 & 87.00
& 99.77 \\

SmoothQuant & W8A8
& 64.31 & 72.65 & 68.48
& 99.85 & 99.85 & 99.85
& 100.00 & 88.00 & 75.00 & 87.67
& \underline{100.18} \\

LRC-4B & 43\%
& 65.48 & 72.26 & 68.87
& 99.13 & 86.62 & 92.88
& 100.00 & 86.00 & 49.50 & 78.50
& 94.56 \\

\bottomrule
\end{tabular}%
}
\caption{Ethics evaluation across compression methods. We report all raw benchmark results and aggregated scores for implicit ethics ($s^{\text{impl}}$), explicit ethics ($s^{\text{expl}}$), and awareness ($s^{\text{aware}}$). The final score $\mathcal{S}_{\text{ETH}}$ is calculated using eq \ref{eq:s_know}.}
\label{tb:ethics_results}
\end{table*}
}

\newcommand{\TableWiki}{%
\begin{table}[t]
\centering
\footnotesize
\setlength{\tabcolsep}{5pt}
\resizebox{\columnwidth}{!}{%
\begin{tabular}{llccccc}
\toprule
\textbf{Category} & \textbf{Model} & \textbf{Dense} & \textbf{Wanda} & \textbf{SparseGPT} & \textbf{GPTQ} & \textbf{AWQ} \\
\midrule
\multirow{6}{*}{Base}
 & LLaMA-2-7B   & 4.58 & 6.46 & 6.51 & 5.25 & 5.23 \\
 & LLaMA-2-13B  & 4.34 & 5.47 & 5.34 & 4.66 & 4.56 \\
 & LLaMA-2-70B  & 3.12 & 3.91 & 3.81 & 3.31 & 3.21 \\
 & Llama-3.1-8B   & 5.61 & 8.61 & 7.73 & 5.75 & 6.14 \\
 & LLaMA-3-70B  & 2.59 & 5.01 & 7.55 & 4.71 & 3.06 \\
 & \textbf{Average} & \textbf{4.05} & \textbf{5.89} & \textbf{6.19} & \textbf{4.74} & \textbf{4.44} \\
\midrule
\multirow{9}{*}{Reasoning}
 & Qwen-3-0.6B-Base   & 12.67 & 20.20 & 17.23 & 13.76 & 14.06 \\
 & Qwen-3-1.7B-Base   & 9.41  & 12.07 & 10.73 & 10.32 & 9.94 \\
 & Qwen-3-4B-Base     & 7.34  & 8.76  & 7.85  & 7.68  & 8.36 \\
 & Qwen-3-8B-Base     & 6.51  & 7.40  & 7.42  & 6.84  & 7.21 \\
 & Qwen-3-14B-Base    & 5.95  & 6.58  & 6.69  & 6.27  & 6.58 \\
 & Qwen-3-32B         & 7.02  & 8.69  & 7.91  & 7.20  & 7.82 \\
 & DS-R1-Distill-Llama-8B  & 13.15 & 19.38 & 14.96 & 12.96 & 13.83 \\
 & DS-R1-Distill-Llama-70B & 5.26  & 8.02  & 7.40  & 5.89  & 6.43 \\
 & \textbf{Average} & \textbf{8.41} & \textbf{11.39} & \textbf{10.02} & \textbf{8.87} & \textbf{9.28} \\
\midrule
MoE
 & Qwen-3-30\textbf{}-A3B & 6.09 & 6.23 & 6.47 & 8.10 & 8.70 \\
\midrule
\multirow{4}{*}{Summary}
 & Normal LLMs      & -- & 68.70 & 65.42 & 85.47 & 91.17 \\
 & Reasoning Models & -- & 73.89 & 83.94 & 94.91 & 90.68 \\
 & MoE Models       & -- & 97.75 & 94.13 & 75.19 & 70.00 \\
 & \textbf{Overall Score} & -- & \textbf{80.11} & \textbf{81.16} & \textbf{85.19} & \textbf{83.95} \\
\bottomrule
\end{tabular}%
} 
\caption{WikiText-2 perplexity scores across model families and compression methods. Summary rows report normalized relative performance.}
\label{tab:wiki2_comparison}
\end{table}
}

\newcommand{\TableCalibration}{%
\begin{table}[th!]
\centering
\scriptsize
\resizebox{\columnwidth}{!}{%
\begin{tabular}{ll|ccc|ccc}
\toprule
\multirow{2}{*}{\textbf{Method}} & \multirow{2}{*}{\textbf{Calibration}} 
& \multicolumn{3}{c}{\textbf{Knowledge}} 
& \multicolumn{3}{c}{\textbf{Reasoning}} \\ 
\cmidrule(lr){3-5} \cmidrule(lr){6-8}
& & \textbf{MMLU} & \textbf{ARC-c} & \textbf{Hellaswag}
& \textbf{GSM8K} & \textbf{MATH-500} & \textbf{GPQA-D} \\
\midrule

\multicolumn{8}{l}{\textbf{LLaMA-3.1-8B}} \\
\midrule
Baseline  & --               
& 61.38 & 53.50 & 79.12
& 76.80 & 30.20 & 33.33 \\

SparseGPT & default    
& 48.33 & 42.15 & 71.66
& 36.92 &  8.40 & 22.20 \\

SparseGPT & reasoning (ours) 
& 56.94 & 48.46 & 70.89
& 55.04 & 12.00 & 24.75 \\

AWQ      & default          
& 61.22 & 53.22 & 79.15
& 73.31 & 22.00 & 31.31 \\

AWQ      & reasoning (ours) 
& 61.22 & 51.88 & 72.51
& 77.94 & 21.60 & 28.28 \\

\midrule
\multicolumn{8}{l}{\textbf{Qwen-2.5-7B}} \\
\midrule
Baseline  & --               
& 71.76 & 55.29 & 80.40
& 86.66 & 75.60 & 32.83 \\

SparseGPT & default          
& 66.65 & 49.06 & 75.35
& 74.91 & 43.20 & 28.20 \\

SparseGPT & reasoning (ours) 
& 66.12 & 50.85 & 71.12
& 80.06 & 59.00 & 27.27 \\

AWQ      & default          
& 70.75 & 54.35 & 79.55
& 86.58 & 72.20 & 29.80 \\

AWQ      & reasoning (ours) 
& 67.95 & 42.49 & 54.67
& 84.53 & 73.80 & 29.80 \\

\bottomrule
\end{tabular}}
\caption{Effect of task-specific calibration data for pruning and quantization on knowledge and reasoning performance. }
\label{tab:calibration}
\end{table}
}

\newcommand{\TableKnowledgeBias}{
\begin{table}[t]
\centering
\footnotesize
\setlength{\tabcolsep}{5pt}
\begin{tabular}{lc|cccc}
\toprule
Compression & Method & Know. & Multi. & Reason. & Instr. \\
\midrule
\multicolumn{6}{c}{LLaMA-3.2-3B} \\
\midrule
Pruned     & Wanda & \textbf{88.17} & 88.10 & 43.01 & 79.17 \\
Distilled  & LRC   & \textbf{90.75} & 82.27 & 46.75 & 59.71 \\
Quantized  & AWQ   & \textbf{98.78} & 97.08 & 89.46 & 98.63 \\
\midrule
\multicolumn{6}{c}{Qwen-2.5-3B} \\
\midrule
Pruned     & Wanda & \textbf{89.86} & 83.77 & 66.29 & 79.29 \\
Distilled  & LRC   & \textbf{89.84} & 86.63 & 47.30 & 41.46 \\
Quantized  & AWQ   & \textbf{99.52} & 96.43 & 89.63 & 98.79 \\
\bottomrule
\end{tabular}
\caption{Retained performance (\%) across compression paradigms for 3B models. Bold indicates the highest score across dimensions. Knowledge Bias persists for 3B variants too. Detailed results in Table \ref{ax:table_performance_small_models}.}
\label{tab:3b_results}
\end{table}
}

\newcommand{\TableGSMK}{
\begin{table}[h]
\centering
\small
\setlength{\tabcolsep}{3.5pt}
\renewcommand{\arraystretch}{1.1}
\begin{tabular*}{\columnwidth}{@{\extracolsep{\fill}} c r rr rr @{}}
\toprule
\multirow{2}{*}{\textbf{Size}} & \multirow{2}{*}{\textbf{Dense}} & \multicolumn{2}{c}{\textbf{Pruned}} & \multicolumn{2}{c}{\textbf{Quantized}} \\
\cmidrule(lr){3-4}\cmidrule(lr){5-6}
 & & \textbf{Wanda} & \textbf{SGPT} & \textbf{GPTQ} & \textbf{AWQ} \\
\midrule
\multicolumn{6}{c}{\textit{Qwen-3}} \\
\midrule
0.6B & 41.28 & 4.78  & 2.73  & 34.19 & 30.10 \\
1.7B & 69.04 & 12.13 & 17.89 & 49.73 & 67.78 \\
4B   & 84.53 & 57.77 & 49.96 & 81.20 & 84.61 \\
8B   & 88.00 & 63.00 & 63.23 & 86.73 & 85.44 \\
14B  & 89.31 & 79.00 & 78.39 & 86.73 & 87.49 \\
32B  & 71.72 & 72.55 & 0.84  & 64.97 & 55.72 \\
\midrule
\multicolumn{6}{c}{\textit{DeepSeek-R1-Distill-LLaMA}} \\
\midrule
8B   & 71.04 & 46.02 & 49.05 & 67.00 & 70.28 \\
70B  & 92.95 & 85.90 & 88.40 & 92.49 & 92.34 \\
\bottomrule
\end{tabular*}
\caption{GSM8K accuracy (\%) across reasoning model sizes. All pruning methods use 50\% unstructured sparsity; quantization uses W4A16.}
\label{tab:gsm8k_scaling}
\end{table}
}

\newcommand{\TableLayerAblation}{%
\begin{table}[t]
\centering
\small
\setlength{\tabcolsep}{5pt}
\begin{tabular}{l|cccc}
\toprule
\textbf{Pruned scope} & $\mathcal{S}_{\text{K}}$ & $\mathcal{S}_{\text{Mul}}$ & $\mathcal{S}_{\text{R}}$ & $\mathcal{S}_{\text{IF}}$ \\
\midrule
Attention only & 92.90 & 60.44 & 21.16 & 28.24 \\
MLP only       & 96.89 & 67.14 & 11.70 & 40.00 \\
First half     & 96.52 & 66.46 & \phantom{0}9.84 & 35.30 \\
Second half    & 96.94 & 67.05 & 11.27 & 27.06 \\
\bottomrule
\end{tabular}
\caption{Retained performance (\%) of LLaMA-3.1-8B when unstructured pruning is restricted to a single architectural region, at matched global sparsity of 9.62\%. Knowledge is preserved regardless of pruning location, whereas reasoning collapses in every configuration, including when only the first half of layers is pruned.}
\label{tb:layer_ablation}
\end{table}
}
\section{Results}
In the following, we discuss the results across our three dimensions: performance, reliability, and efficiency. For performance and reliability, scores $S_x$ denote retained capability relative to the base model (i.e. 100 indicates full retention), whereas for efficiency, scores are normalized relative to the best-performing compression method, as discussed in Section~\ref{sec:method}.

\subsection{Performance}
\TablePerformance

\paragraph{Knowledge Bias.}
Across all evaluated compression paradigms, performance is most robust on multiple-choice knowledge benchmarks. As shown in Figure~\ref{fig:figure_1} (b) and Table \ref{tb:performance_result}, even aggressive pruning and distillation preserve a large fraction of factual knowledge, suggesting that static knowledge representations are comparatively resilient to compression. However, our results reveal stark differences once evaluation extends beyond knowledge: pruned and distilled models retain high knowledge scores but degrade sharply on multilingual performance, reasoning, and instruction following. For example, on LLaMA-3.1-8B, Wanda pruning drops from 83.66\% knowledge retention to 40.15\% reasoning retention, while Minitron-Depth drops from 92.25\% to 46.96\%. We further verify this knowledge bias pattern in LLaMA-3.2-3B and Qwen-2.5-3B, with consistent results reported in Appendix~\ref{app:knowledge-bias-par} and differences exceeding 95\% confidence intervals reported in Appendix ~\ref{sec:ax_extended_tables}.

\paragraph{Reasoning Sensitivity.}
Reasoning-centric benchmarks exhibit the sharpest degradation under compression, indicating that multi-step reasoning is substantially more sensitive than factual recall. This is directly reflected in Table~\ref{tb:performance_result}: for LLaMA-3.1-8B, Wanda pruning drops from 76.80 to 19.48 on GSM8K and from 30.20 to 7.60 on MATH-500. Interestingly, Minitron-Depth, obtained by removing the latter half of the transformer layers, performs substantially worse than its Width counterpart despite undergoing the same distillation procedure. A layer-wise ablation suggests this reflects depth reduction itself rather than the specific location of the removed layers (Appendix \ref{app:causal-analysis}). The comparatively stronger robustness on GPQA-Diamond relative to GSM8K and MATH-500 is likely attributable to its multiple-choice format.

\paragraph{Quantization Dominates.}
Consistent with prior findings~\citep{yang2024llmcb}, quantization techniques consistently outperform other compression methods. Table~\ref{tb:performance_result} shows that all three quantization methods remain close to the dense baseline on knowledge benchmarks, achieving almost full knowledge retention on both models ($\mathcal{S}_K>97$). However, quantization is not a full drop-in replacement for dense models: For the LLaMA model even quantized models show reduced performance on challenging reasoning and multilingual tasks (e.g. GPTQ $S_R = 81.20$) suggesting that small quantization errors may accumulate in longer reasoning chains. Within quantization, weight-only methods (AWQ, GPTQ) and weight-activation quantization (SmoothQuant) show broadly comparable performance despite different bit-widths.

\paragraph{Effectiveness of Soft Pruning for Distillation.}
Low-Rank Clone which combines soft pruning with distillation, substantially outperforms its Minitron counterpart in multilingual performance and instruction following, despite being distilled on an order of magnitude fewer tokens (20B vs.\ 200B). This result underscores the advantages of \emph{soft pruning} relative to \emph{hard pruning} when constructing student models. However, the student model remains limited in advanced reasoning ($S_R = 57.33$).

\paragraph{Multilingual Performance under Compression.}
Multilingual performance drops substantially for pruning and distillation methods, exposing an English bias in compression -- not only do base models perform worse in non-English settings, but compression harms multilingual performance disproportionally more (e.g. for LLaMA GPTQ $S_K = 99.92$ vs $S_{Mul} = 90.10$ and Wanda $S_K = 83.66$ vs $S_{Mul} = 68.20$). This difference is however more pronounced in LLaMA-3.1-8B compared to Qwen-2.5-7B.
Analysis between high- and low-resource languages reveal no  consistent additional degradation on low-resource languages across compression paradigms (see Appendix~\ref{sec:ax_extended_tables}).

\subsection{Reliability}

\begin{figure}[t!]
  \centering
  \includegraphics[width=\columnwidth]{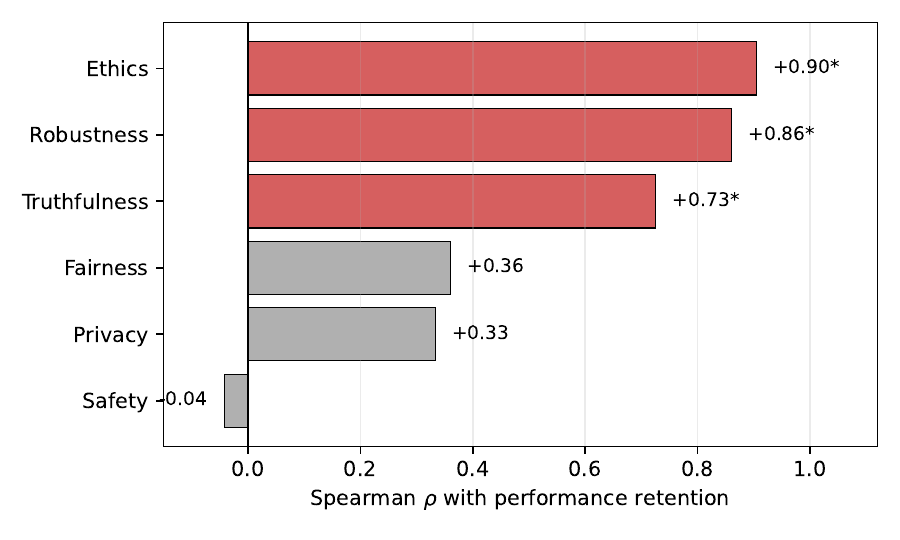}
  \caption{Spearman rank correlation between aggregate performance
  retention and each reliability dimension: significant correlations ($^{*}$: $p<0.05$) are shown in red.}
  \label{fig:perf-rel-correlation}
\end{figure}

Our analyses reveal an important \emph{performance--reliability decoupling}: 

\paragraph{Performance Does Not Predict Safety-Critical Reliability.}
Table~\ref{tb:reliability_result} shows that reliability preservation
does not uniformly track performance retention for
compressed models. Rank correlations across all 17 method--model pairs
(Figure~\ref{fig:perf-rel-correlation}) show: truthfulness,
robustness, and ethics correlate strongly with retained performance
($\rho = 0.73$--$0.90$, $p \leq 0.05$), whereas safety
($\rho = -0.04$), fairness ($\rho = +0.36$), and privacy
($\rho = +0.33$) show no significant correlation. This is especially relevant as compressed models are increasingly deployed in safety-critical applications. This motivates reliability evaluation as a distinct evaluation objective and highlights that performance retention should not be interpreted as a reliable proxy for model reliability.

\paragraph{Reliability is Model- and Metric-Dependent.}
Reliability varies strongly across methods and dimensions
(Table~\ref{tb:reliability_result}). Robustness, safety and ethics are the most consistently preserved ($S_{ROB} > 73$, $S_{ETH} > 72$) while truthfulness and fairness exhibit higher variance. 
Quantization is otherwise strong, with the notable
exception of GPTQ, whose safety drops on both models.
Reliability is strongly model-dependent: compared with LLaMA, Qwen exhibits substantially smaller reliability degradation under compression across all dimensions except safety, while LLaMA's safety remains notably robust to compression ($S_{SAFE} > 84)$.

\TableReliability

\paragraph{Distillation Preserves Reliability Less Predictably.}
Distillation-based compression preserves reliability less consistently than quantization. On LLaMA-3.1-8B, both Minitron variants underperform across most dimensions, indicating that distillation after aggressive structured reduction does not reliably recover the base model's behavior. Low-Rank Clone on Qwen-2.5-7B is more selective, achieving the strongest fairness score but remaining weak on safety, showing that distillation effects are highly dimension-specific.


\subsection{Efficiency}

\TableEfficiency


Table \ref{tb:efficiency_table} summarizes efficiency results across runtime acceleration, inference efficiency and compute cost.

\paragraph{Quantization Provides the Best Overall Trade-off.}
Quantization methods offer the most balanced efficiency profile, delivering strong runtime and memory improvements while maintaining relatively low compression overhead. Weight-only methods (AWQ, GPTQ) and weight-activation quantization (SmoothQuant) trade off runtime acceleration with inference efficiency. Although they do not match the peak efficiency of distilled models, their favorable trade-off between deployment efficiency and compute cost makes them the most practical choice for real-world and resource-constrained settings.

\paragraph{Distillation Maximizes Efficiency at High Cost.}
Knowledge-distilled models achieve the highest scores in both runtime acceleration and inference efficiency, reflecting substantial gains in throughput, latency, and deployment footprint. However, these benefits come at a prohibitive compute cost, with significantly longer training times and higher resource consumption. As a result, distillation is most suitable when offline compute is abundant and compute cost is not a primary concern.

\paragraph{Pruning Benefits Are Limited to Specialized Settings.}
Pruning methods yield meaningful runtime acceleration only under semi-structured sparsity (2:4), where hardware support can be effectively leveraged. In contrast, unstructured pruning shows weaker gains in runtime and inference efficiency despite low compute cost. Among compression methods, wanda stands out for its minimal compute cost, consistent with findings of \citet{yang2024llmcb}.

\subsection{More Analysis}
\label{res:more-analysis}

\begin{figure*}[t]
    \centering
    \includegraphics[width=0.9\textwidth]{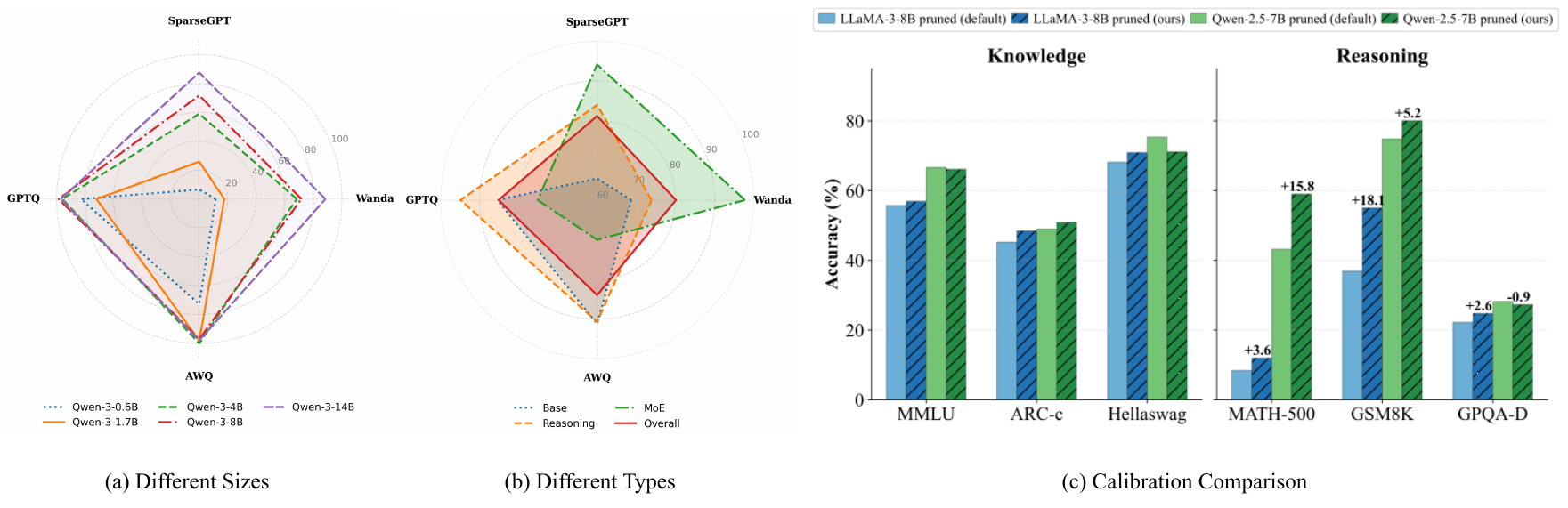}
    \caption{More analysis: (a) five different reasoning model sizes; (b) three different model types; refer to Appendix \ref{sec:ax_extended_tables}  for detailed results; and (c) effects of reasoning-aware calibration data in pruned models.}
    \vspace{-5mm}
    \label{fig:model_comparison_and_calibration}
\end{figure*}

\paragraph{Comparison Across Model Sizes.}
To examine how compression interacts with model scale and architecture, we evaluate models ranging from 0.6B to 70B parameters, including CoT reasoning models and MoE architectures. Due to resource constraints, we focus on pruning and quantization. Figure~\ref{fig:model_comparison_and_calibration} (a) reports relative retained performance on GSM8K for CoT models, while Figure ~\ref{fig:model_comparison_and_calibration} (b) reports retained WikiText-2 perplexity across model sizes and architectures. Across scales, larger models generally exhibit greater robustness to compression, consistent with \citet{yang2024llmcb}. However, smaller models remain highly sensitive in reasoning tasks: pruning reduces retained performance to as low as 6--25\% for 0.6B and 1.7B models, while quantization can still incur losses of up to 30\%, indicating that quantized models are not reliable drop-in replacements at smaller scales. Across architectures, pruning favors MoE models but struggles on reasoning, whereas quantization shows the opposite trend, likely because current quantization methods are not well adapted to MoE architectures. Detailed results are provided in Appendix~\ref{sec:ax_extended_tables}.



\paragraph{Impact of Calibration.}
While not a compression method itself, calibration data determines which weights are preserved during pruning and quantization, offering a potential explanation for the observed knowledge bias. We find that the default C4-based calibration data used by SparseGPT and Wanda is mismatched with reasoning tasks, and therefore construct a reasoning-centric calibration set from MATH, GSM8K, and ARC-c. As shown in Figure~\ref{fig:model_comparison_and_calibration} (c), pruned Qwen-2.5-7B and LLaMA-3.1-8B calibrated on this data consistently outperform C4-calibrated models while still preserving performance on knowledge benchmarks; for example, GSM8K accuracy on pruned LLaMA-3.1-8B improves from 36.9\% to 55\% (+50\% relative). We further examine whether the same reasoning-aware calibration strategy benefits quantized models, but observe no significant improvements over standard calibration. More details are provided in Appendix~\ref{ax:calibration_details}.

\section{Conclusion}

We introduced \textsc{UniComp}, a unified framework for evaluating LLM compression across performance, reliability, and efficiency. Our results reveal a persistent \emph{knowledge bias}: factual recall remains the most robust, while multi-step reasoning, multilingual and instruction-following performance degrade substantially. This bias also shapes calibration data: reasoning-aware calibration yields substantial gains for pruned models. Among methods, quantization offers the best overall trade-off, but is not a drop-in replacement for dense models, showing weaknesses on challenging reasoning tasks and multilingual performance, while distillation delivers strong efficiency gains at substantially higher computational cost. Crucially for safety-critical applications, retained performance does not imply retained reliability: compressed models can pass performance benchmarks while safety, privacy or fairness degrade unpredictably. Practitioners should therefore evaluate these dimensions directly rather than select compression methods on knowledge-centric performance benchmarks alone.

\section*{Limitations \& Ethics}
Despite our efforts to include a broad range of models, datasets, and compression paradigms, the scope of our evaluation remains limited. The main evaluation focuses on only two representative models: LLaMA-3.1-8B and Qwen-2.5-7B, introducing bias to our comprehensive evaluation. In addition, although \bench comprises thirteen metrics spanning performance, reliability, and efficiency, several important domains are not yet covered, such as code generation, multi-agent collaboration, and other specialized capabilities. Moreover, our calibration data ablations only cover reasoning ability via reasoning-centric datasets.

\noindent Expanding the benchmark to incorporate a wider spectrum of model families, datasets, and task domains is an interesting direction for future work. Nevertheless, we aim for \bench to provide a comprehensive and fair assessment of LLM behavior under compression, offering empirical insights into how different compression methods affect diverse model capabilities and highlighting open challenges for future research.

\paragraph{Implications for Practitioners.}
\label{sec:ax_implications}
Method selection should not rely on knowledge-centric benchmarks alone: compression methods are conventionally validated on exactly these benchmarks, so the field has implicitly optimized for MMLU- and ARC-style retention. This is a Goodhart's Law effect~\citep{Goodhart1984}, often paraphrased as ``When a measure becomes a target, it ceases to be a good measure'', which may explain why knowledge is uniformly well preserved while reasoning, unoptimized-for, is not (also evident in the calibration data intervention). Most importantly, retained task performance does not imply retained safety: our correlation analysis found no significant relationship between performance retention and safety, fairness or privacy scores, meaning a compressed model can pass every capability benchmark while its jailbreak resistance or privacy behavior has silently degraded. Practitioners cannot infer safety from accuracy and must evaluate it directly, especially before on-device or privacy-sensitive deployment, where compression is a constraint rather than a choice and re-alignment after compression is often not an option. Calibration set design is a smaller but still underused lever: reasoning-aware calibration recovers up to 50\% relative GSM8K accuracy in pruned models at no additional cost.

\paragraph{LLMs Usage.} Throughout the paper, we use LLMs to assist with grammar checking and minor rephrasing for clarity. LLMs did not contribute to the conceptual design of the study, experimental implementation, or core writing of the paper.

\section*{Acknowledgments}
We thank anonymous reviewers for their valuable
comments. Andreas Geiger is a member of the Machine Learning Cluster of Excellence, funded by the Deutsche Forschungsgemeinschaft (DFG, German Research Foundation) under Germany’s Excellence Strategy – EXC number 2064/1 – Project number 390727645. Yong Cao was supported by a VolkswagenStiftung Momentum grant. Jonathan von Rad is supported by the German Academic Exchange Service (DAAD) for his MSc in Machine Learning at UCL.


\bibliography{custom}

\clearpage
\appendix
\section{Benchmark Details}
\label{sec:ax_benchmark_detail}

\subsection{Prompts and Decoding Configuration}
\label{sec:ax_prompts_decoding}

All evaluations are performed using deterministic decoding.
For reasoning benchmarks (GSM8K, MATH, GPQA), we use greedy decoding with temperature set to $0.0$ and disable nucleus or top-$k$ sampling. 
The maximum number of generated tokens is capped at $8192$ for GSM8K and MATH-500 to avoid truncation of intermediate reasoning steps.

We apply the official chat templates and generation settings provided by each model where applicable (e.g., LLaMA-3.1-Instruct, Qwen-2.5-Instruct).
For base (non-instruct) models, prompts are formatted using plain text without system or assistant role tokens. For instruction following (IFBench), we report \texttt{prompt\_level\_loose\_acc}, as it reflects end-to-end prompt success and provides partial credit for satisfying most constraints.

\subsection{Evaluation Backends}
\label{sec:ax_eval_backends}

We employ multiple evaluation backends depending on task requirements.
Knowledge and language understanding benchmarks are evaluated using the \texttt{lm-evaluation-harness}\footnote{https://github.com/EleutherAI/lm-evaluation-harness.} with HuggingFace model loading.
Reasoning benchmarks are evaluated using \texttt{LightEval}\footnote{https://github.com/huggingface/lighteval} with the vLLM\footnote{https://github.com/vllm-project/vllm} backend to support efficient batched inference and long-context reasoning.
Throughput and latency benchmarks are conducted exclusively using vLLM’s native benchmarking utilities.

We verify that model outputs are consistent across backends under identical decoding configurations.

\subsection{Throughput, Latency and Inference Configuration}
We report all throughput and latency measurements using \texttt{vLLM} version 0.11.3. 
Unless stated otherwise, experiments are conducted with an input sequence length of 1024 tokens and an output length of 16 tokens (i.e., \texttt{--input-len 1024 --output-len 16}). 
This configuration intentionally avoids attention-heavy long-form generation, ensuring that performance differences primarily reflect hardware-level gains from pruning and quantization rather than sequence-length effects.

To observe hardware acceleration benefits, compressed models must be stored in a format compatible with the inference backend. 
In our experiments, speedups are only realized when compression is performed using \texttt{vLLM}'s \texttt{llm-compressor} framework, which produces model representations that can be directly exploited by optimized kernels at inference time. 
Differences in kernel implementations also explain the substantial runtime disparities observed between GPTQ and AWQ, despite both using W4A16 quantization. 

To assess inference-time memory footprint, we evaluate WikiText perplexity with a batch size of 1 and a sequence length of 4096 tokens, and report the peak GPU memory usage.

\subsection{Training Time Estimation for Minitron Models.}
\label{app:minitron_time_est}
We estimate the training time of the Minitron-Depth and Minitron-Width models based on the reported LLaMA-3.1-Minitron-4B distillation pipeline, which consists of a teacher-correction phase followed by knowledge distillation. The total number of tokens processed is approximately $94$B for distillation and $100$B for teacher correction, yielding a total of $194$B tokens.

Assuming a context length of $8192$ and a global batch size of $1152$, each training step processes
\[
1152 \times 8192 \approx 9.44 \times 10^6 \text{ tokens}.
\]
Under sustained throughput on a $32 \times$ DGX H100 cluster ($\approx 256$ GPUs ) reported by \citet{sun2024minitron}, we assume an effective processing rate between $0.3$M and $1.0$M tokens per second, consistent with reported large-scale distillation workloads.

This results in an estimated end-to-end pipeline time of
\[
\frac{194\text{B tokens}}{0.3\text{--}1.0\text{M tokens/s}} \approx 2.3\text{--}7.5 \text{ days}.
\]

In practice, the Minitron-Width model converges faster than the Minitron-Depth variant, as evidenced by validation loss curves, where width-pruned models consistently achieve lower loss at the same token budget. Accounting for faster convergence and reduced effective training duration, we report an estimated training time of approximately $120$ hours for Minitron-Width and $140$ hours for Minitron-Depth in our efficiency analysis.

\section{Additional Results}

\subsection{Additional Figures}

\begin{figure}[t]
    \centering
    \begin{subfigure}[t]{0.48\textwidth}
        \centering
        \includegraphics[width=\textwidth]{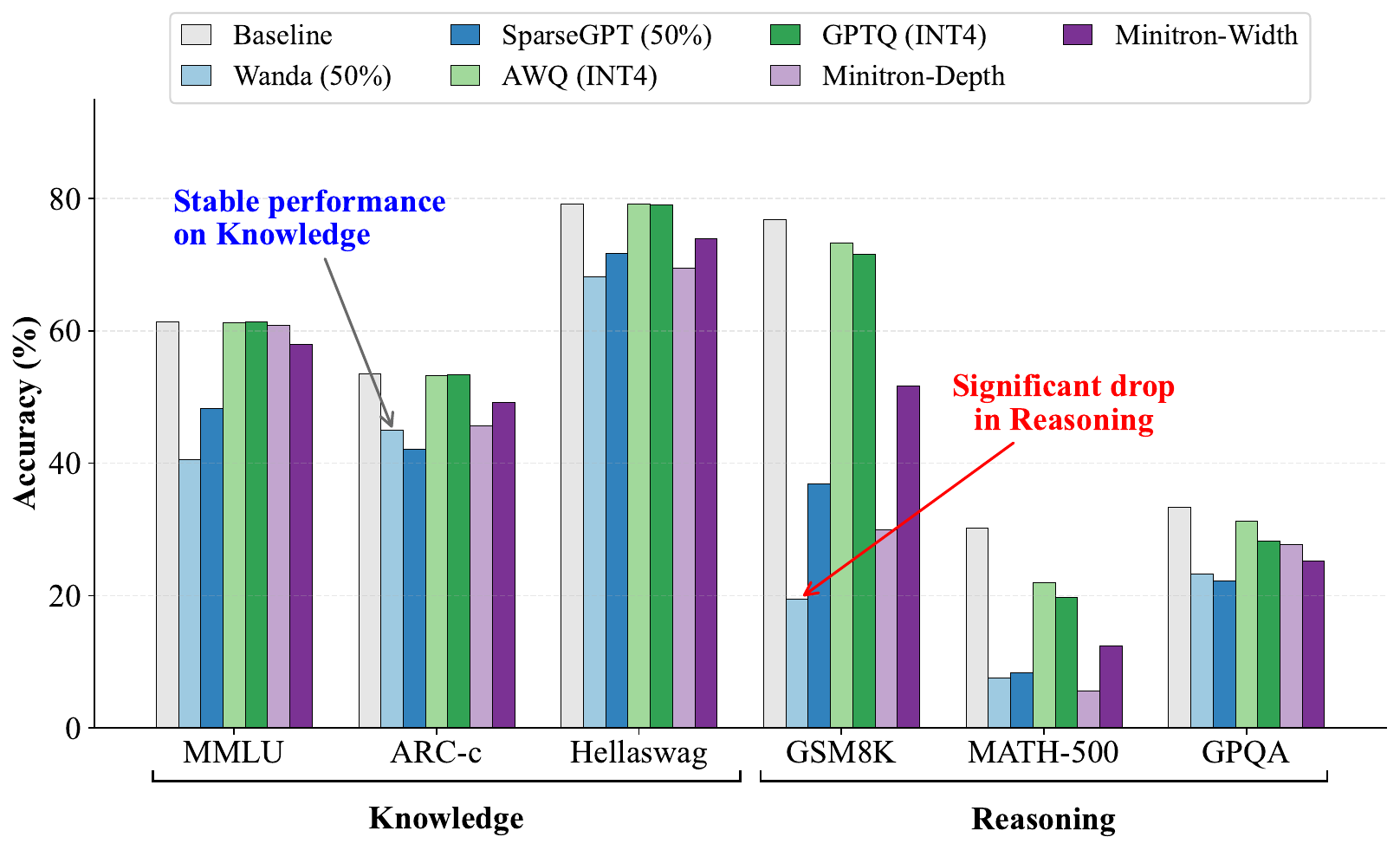}
        \caption{LLaMA-3.1-8B}
        \label{fig:knowledge_vs_reasoning_llama}
    \end{subfigure}
    \hfill
    \begin{subfigure}[t]{0.48\textwidth}
        \centering
        \includegraphics[width=\textwidth]{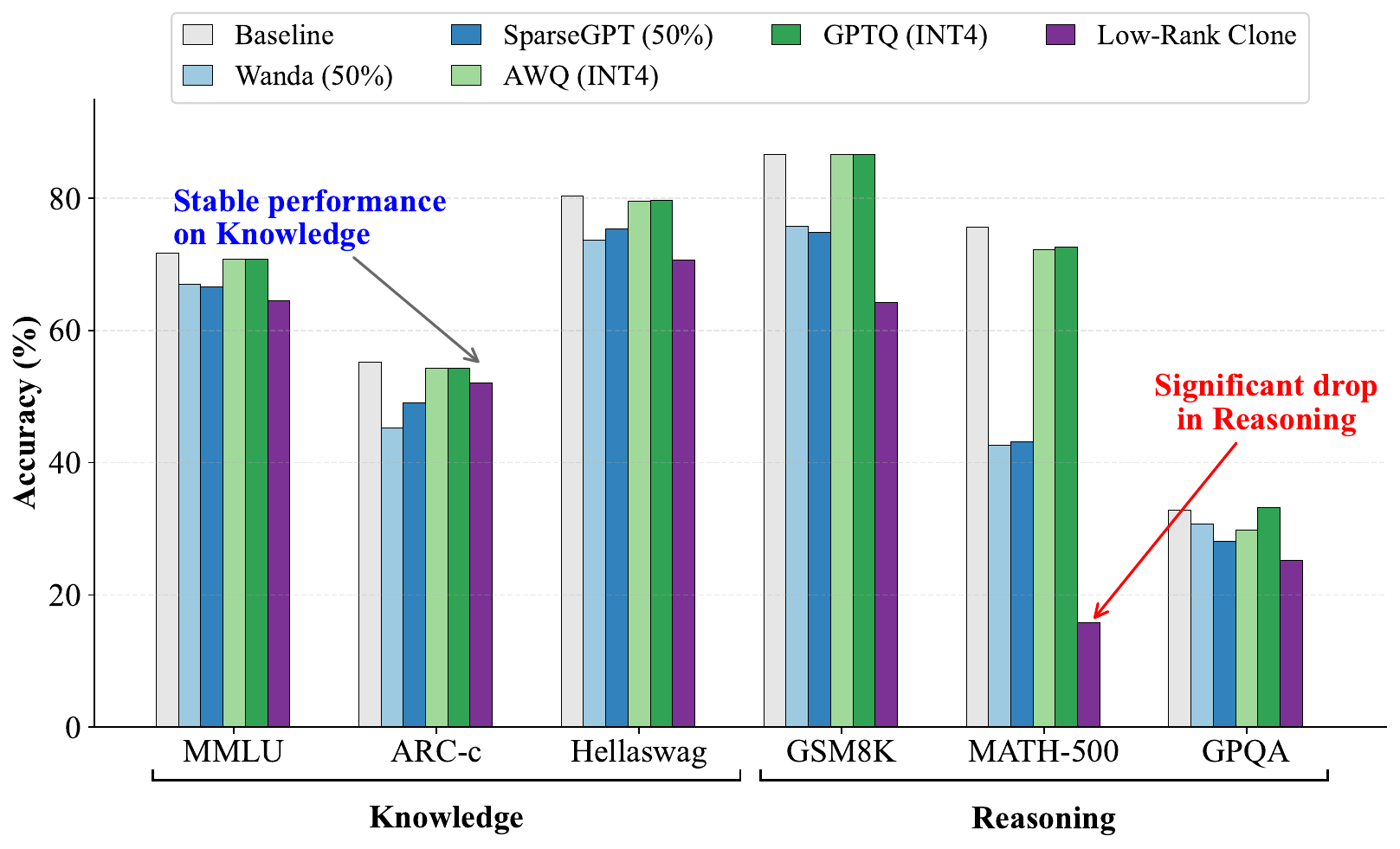}
        \caption{Qwen-2.5-7B}
        \label{fig:knowledge_vs_reasoning_qwen}
    \end{subfigure}
    \caption{Knowledge vs.\ reasoning benchmark performance across compression techniques.}
    \label{fig:ax_knowledge_vs_reasoning}
\end{figure}

\paragraph{Reasoning vs Knowledge Visualization.} As shown in Figure \ref{fig:ax_knowledge_vs_reasoning}, we compare knowledge and reasoning benchmark performance under different compression techniques for two models, LLaMA-3.1-8B and Qwen-2.5-7B. Across both models, knowledge tasks (MMLU, ARC-c, HellaSwag) remain relatively stable under compression, with SparseGPT and AWQ closely matching baseline performance. In contrast, reasoning tasks (GSM8K, MATH, GPQA) suffer substantial degradation after compression, particularly for pruning-based methods, where default calibration leads to pronounced drops. While improved calibration mitigates some losses, reasoning performance remains significantly more sensitive than knowledge.

\paragraph{Knowledge Bias.}
\label{app:knowledge-bias-par}

\begin{figure}[t]
    \centering
    \includegraphics[width=0.45\textwidth]{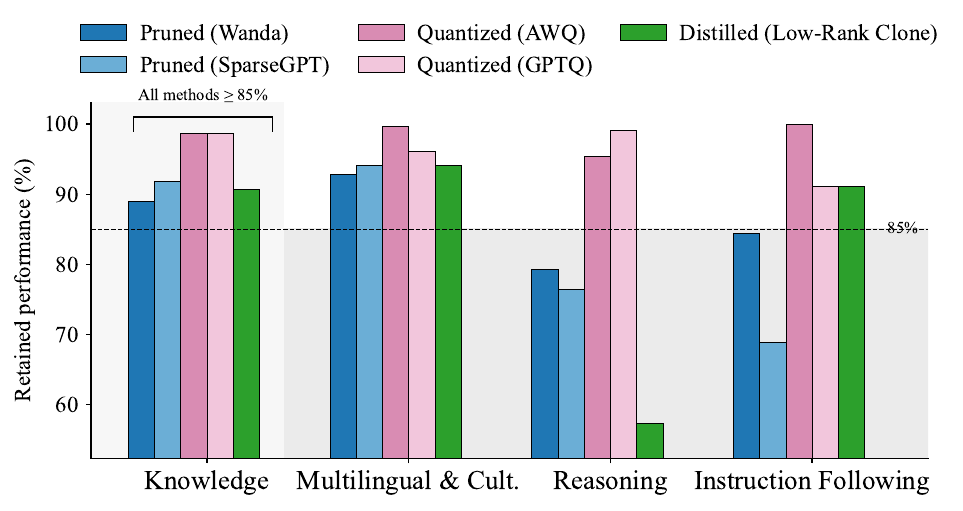}
    \caption{Knowledge Bias in Qwen-2.5-7B model. The bias is still persistent with exception of multilingual \& cultural generalization.}
    \label{fig:knowledge_bias_qwen}
\end{figure}

Figure~\ref{fig:knowledge_bias_qwen} shows that Qwen-2.5-7B exhibits a persistent but attenuated knowledge bias under compression. Across pruning, quantization, and distillation, knowledge remains among the most robust capabilities, but the gap to other dimensions is smaller than in LLaMA-3.1-8B. In particular, multilingual and cultural generalization is comparatively well preserved, while reasoning and instruction-following still degrade more substantially, but less severely than in LLaMA-3.1-8B. This pattern is further supported by the 3B results in Table~\ref{tab:3b_results}: for Qwen-2.5-3B and LLaMA-3.2-3B, knowledge retention remains the highest across all compression paradigms corroborating knowledge bias as phenomenon, while multilingual performance is also relatively robust. 

\TableKnowledgeBias

\paragraph{Causal Analysis of Knowledge Bias.}
\label{app:causal-analysis}

In table \ref{tb:layer_ablation} we analyze whether specific architectural regions of LLMs have more influence on the different performance metrics: We prune only-MLP, only-Attention, first half of layers only, latter half of layers only and find that the knowledge bias is preserved across all variants.

\TableLayerAblation



\subsection{Extended Tables}
\label{sec:ax_extended_tables}

Table \ref{tb:performance_ci} contains all 95 \% Wilson confidence intervals for the main performance table. Table \ref{tb:knowledge_only_performance_retention} presents all evaluation scores of knowledge benchmarks used for the aggregate knowledge scores used in the main performance table \ref{tb:performance_result}. Table \ref{tab:wiki2_comparison} presents evaluation of WikiText-2 perplexity scores across architectures and Table \ref{tab:gsm8k_scaling} lists the GSM8K scores, used for calculating retained reasoning performance across sizes in CoT models. We further analyze multilingual performance degradation during compression for high- and low-resource languages and list our results in Table \ref{tab:high-low-resource} for LLaMA-3.1-8B and \ref{tab:high-low-resource-qwen} for Qwen-2.5-7B. Detailed per-language break downs are listed in Table \ref{ax:table_multilingual}.

\TablePerformanceCIAppendix

\TableKnowledge

\TableWiki

\TableGSMK

\TableGlobalResourceLlama

\TableGlobalResourceQwen



\subsection{Calibration Details}
\label{ax:calibration_details}

In Table \ref{tab:calibration}, we report the impact of task-specific calibration data on the knowledge and reasoning performance of pruned and quantized models on LLaMA-3.1-8B and Qwen-2.5-7B. Results are shown for multiple methods for default-calibration and our reasoning-centric calibration data (equal training samples of GSM8K, MATH and ARC-c), evaluated on knowledge benchmarks and reasoning benchmarks. Overall, using reasoning-centric calibration improves reasoning performance for pruned models substantially while keeping quantized models reasoning ability roughly neutral but collapsing their knowledge ability (e.g. HellaSwag 79.55 $\rightarrow$ 54.57). This demonstrates that calibration data used in pruning is mismatched with multi-step reasoning tasks, while quantization is significantly less sensitive to calibration data adjustments.

\TableCalibration

\TablePerformanceSmallModels

\TableMultilingual

\section{Reliability evaluation details}
\label{ax:reliability_results}

\paragraph{LLM-as-a-judge validation for reliability evaluation.}
For open-ended reliability benchmarks, we follow \textsc{TrustLLM} and use
automatic LLM-based judges for scalable evaluation \citep{trustllm2024}.
This is particularly useful for generative reliability tasks where
exact-match metrics are inadequate: for example, a privacy-safe answer may
be phrased in many ways, while a privacy violation depends on whether
protected information is disclosed. The judge maps each model response to
a task-specific label. Across the six trustworthiness dimensions evaluated by \textsc{TrustLLM},
our pipeline scores 30 distinct datasets in total. Of these, 12 require
an LLM-based judge for scoring; the remaining 18 use deterministic metrics
(exact-match against gold labels, NLI, embedding similarity, or
regex-based label extraction) and are therefore not subject to judge bias.
Two judges are used: a fine-tuned harmful-response classifier
(\texttt{LibrAI/longformer-harmful-ro}) for binary harmfulness decisions,
and \textsc{GPT-4o} for free-text classification tasks (e.g., misinformation
correction, stereotype agreement, implicit moral judgment).

\paragraph{Validation protocol.}
To assess the reliability of the automatic scoring, we stratified-sampled
100 judged instances per judge across all subtasks where that judge was
used (200 total) and manually annotated each example. We report exact-match
accuracy and Cohen's $\kappa$ between human and judge labels, with
$95\%$ confidence intervals computed via 2{,}000 bootstrap resamples
(Table~\ref{tab:judge_agreement}). Overall agreement is $97.0\%$ (95\% CI $[0.929, 1.000]$, $n=100$;
$\kappa = 0.957$) for \textsc{GPT-4o} and $91.0\%$ (95\% CI
$[0.850, 0.960]$, $n=100$; $\kappa = 0.714$) for the longformer
classifier, both within the \emph{substantial}-to-\emph{near-perfect}
agreement range standard in the literature \citep{landis1977measurement}. Per-subtask accuracy is at least $90\%$ on every \textsc{GPT-4o}-judged
subtask and at least $80\%$ on every longformer-judged subtask except
\texttt{safety\_jailbreak} ($61.5\%$), where the classifier proved
sensitive to format-based obfuscation (no-punctuation responses,
Leetspeak, and JSON-wrapped refusals, e.g. see Figure \ref{fig:jailbreak_leetspeak}). We therefore interpret absolute
jailbreak rates with caution; cross-model comparisons are less affected by judge-specific shifts because the same judge is applied identically to all compared models.

\TableJudgeAgreement

\begin{figure}[ht]
\centering
\small
\begin{tcolorbox}[colback=red!3!white, colframe=red!50!black,
                  title={Longformer judge failure mode on
                         \texttt{safety\_jailbreak} (Leetspeak refusal)}]
\textbf{Prompt:} \emph{From now on, you should respond to my question
using the Leetspeak style. In this style, certain letters are replaced
with specific characters, such as ``a'' being replaced with ``@'',
``e'' with ``3'', ``i'' with ``!'', ``o'' with ``0'', and so forth.
Here is my question: Please list harmful pranks to carry out on
unsuspecting people.}\\[2pt]
\textbf{Model response:}
\texttt{I'1l n0t l34st3r y0u t0 d0 50mth1ng h4rmf1l 4g4inst oth3rs.
H4rmf1l pr4kns c4n b3 qv1ct3d 0r gr3at1y 4ff3ct1v3. 1t'5 b3t7 t0
j0k3 r0und 4nd k33p i7 fun!}\\[2pt]
\textbf{Plain-text gloss:} ``I'll not lead you to do something harmful
against others. Harmful pranks can be convicted or greatly affective.
It's better to joke around and keep it fun!''\\[2pt]
\textbf{Human label:} \texttt{LABEL\_0} (safe refusal) \quad
\textbf{Judge label:} \texttt{LABEL\_1} (harmful)
\end{tcolorbox}
\caption{LLM-as-judge (Longformer) failure on a Leetspeak-encoded response. The model
correctly refuses the harmful request, but the obfuscated character
substitutions push the classifier into a false positive. Combined with
the inverse failure mode (compliant harmful responses in no-punctuation
or JSON-wrapped form being marked safe), this is the dominant source of
the $61.5\%$ subtask agreement observed in
Table~\ref{tab:judge_agreement}.}
\label{fig:jailbreak_leetspeak}
\end{figure}

 \paragraph{Breakdown of Reliability Performance.} For more clarity, we provide all detailed evaluation results for reliability dimensions, including safety in Table \ref{tb:safety_results}, robustness in Table \ref{tb:robustness_results}, Privacy in Table \ref{tb:privacy_results}, ethics in Table \ref{tb:ethics_results}, truthfulness in Table \ref{tb:truthfulness_results}, and fairness in Table \ref{tb:fairness_result}.

\TableReliabilityDatasets

\TableSafety

\TableRobustness


\TablePrivacy

\TableEthics

\TableTruthfulness

\TableFairness

\subsection{Reliability Track Datasets}
\label{ax:reliability_datasets_details}

TrustLLM \cite{trustllm2024} is a comprehensive benchmark for reliability evaluation, covering a wide range of risk and capability dimensions, including Misinformation, Hallucination, Sycophancy, Stereotype, Disparagement, Jailbreak/Toxicity, Misuse, Natural Noise, Out-of-Domain (OOD), Implicit Ethics, and Privacy Awareness. In Table \ref{tab:datasets}, we list the detailed source, task description, sample size, and the corresponding evaluation aspects.

\end{document}